\documentclass[journal]{vgtc}

\vgtccategory{Research}

\vgtcpapertype{algorithm/technique}


\teaser{
  \centering
  \includegraphics[width=\linewidth]{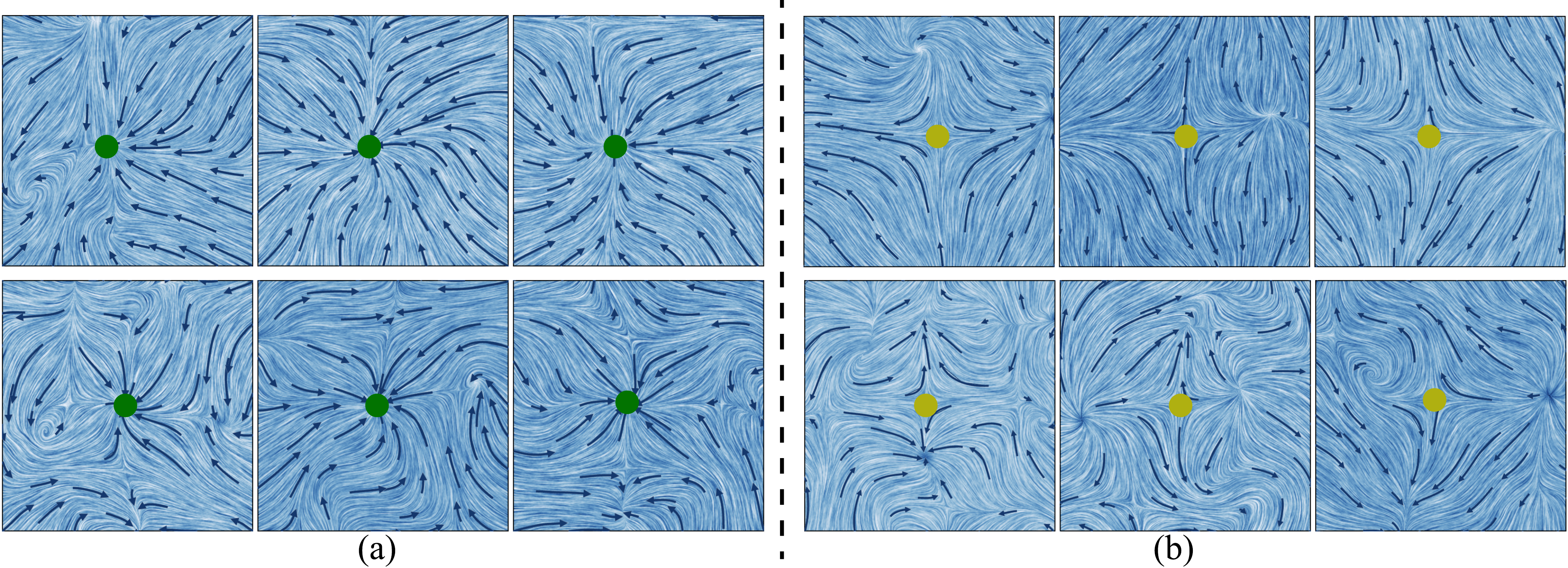}
  \caption{%
We show topology-guided sampling, for different distributions formed over a simulation ensemble. (a) Generating a sink in the area center. (b) Generating a saddle in the area center. The first row shows samples from later timesteps of the simulation, while the second row shows samples from earlier timesteps. Topology-guided generation of vector fields reveals that flow fields exhibit more vortices and complex spatial evolution in the early timesteps compared to the later timesteps, given the same local topology features, e.g. common sinks, common saddles.
  }
  \label{fig:teaser}
}

\graphicspath{{figs/}{figures/}{pictures/}{images/}{./}} 

\usepackage{lipsum}                    
\usepackage{mwe}                       
\usepackage{amsmath,bm}
\usepackage{amssymb}

\usepackage{microtype}                 
\PassOptionsToPackage{warn}{textcomp}  
\usepackage{textcomp}                  
\usepackage{mathptmx}                  
\usepackage{times}                     
\usepackage{cite}                      
\usepackage{tabu}                      
\usepackage{booktabs}                  

\usepackage{algorithm, algcompatible, algpseudocode, amsfonts}
\usepackage{multirow}
\usepackage{mathrsfs}

\DeclareMathOperator{\Tr}{Tr}

\newcommand{\be}{{\bm{\epsilon}}}
\newcommand{\baralpha}{{\sqrt{\bar{\alpha}_t}}}
\newcommand{\baralphaone}{{\sqrt{1-\bar{\alpha}_t}}}

\onlineid{1334}

\vgtccategory{Research}

\title{Topology Guidance: Controlling the Outputs of Generative Models via Vector Field Topology}

\author{
  Xiaohan Wang \\ \scriptsize Vanderbilt University \and
  Matthew Berger\thanks{e-mail: matthew.berger@vanderbilt.edu}\\
  \scriptsize Vanderbilt University
}

\abstract{
For domains that involve numerical simulation, it can be computationally expensive to run an ensemble of simulations spanning a parameter space of interest to a user.
To this end, an attractive surrogate for simulation is the generative modeling of fields produced by an ensemble, allowing one to synthesize fields in a computationally cheap, yet accurate, manner.
However, for the purposes of visual analysis, a limitation of generative models is their lack of control, as it is unclear what one should expect when sampling a field from a model.
In this paper we study how to make generative models of fields more controllable, so that users can specify features of interest, in particular topological features, that they wish to see in the output.
We propose topology guidance, a method for guiding the sampling process of a generative model, specifically a diffusion model, such that a topological description specified as input is satisfied in the generated output.
Central to our method, we couple a coordinate-based neural network used to represent fields, with a diffusion model used for generation.
We show how to use topologically-relevant signals provided by the coordinate-based network to help guide the denoising process of a diffusion model.
This enables us to faithfully represent a user's specified topology, while ensuring that the output field remains within the generative data distribution.
Specifically, we study 2D vector field topology, evaluating our method over an ensemble of fluid flows, where we show that generated vector fields faithfully adhere to the location, and type, of critical points over the spatial domain.
We further show the benefits of our method in aiding the comparison of ensembles, allowing one to explore commonalities and differences in distributions along prescribed topological features.
}

\begin{document}


\firstsection{Introduction}

\maketitle
    
Our understanding of real-world physical phenomena can often be advanced by computational models of our physical surroundings. 
To this end, numerical simulation, e.g. the solving of partial differential equations (PDE), still remains the predominant approach for such modeling and analysis.
Simulations are attractive as they allow us to account for the uncertainty that is inherent in physical phenomena.
This is often achieved by simulating a collection of models that vary over a predefined parameter space, ranging from initial/boundary conditions, to adjustable input parameters.
However, the execution of multiple simulations is notably resource-intensive, presenting a significant barrier due to the high computational costs and I/O bottleneck associated with multiple runs.
Recent work has sought to alleviate these compute \& network burdens by representing an ensemble of simulation outputs with deep generative models~\cite{kohlBenchmarking2024}.
A cheaply-drawn sample from a generative model is intended to represent a plausible output from an otherwise expensive model-based simulation.
Generative models thus have the potential to ``fill in the gaps'' left by the outputs of a simulation ensemble, allowing us to rapidly obtain realistic instances of physical phenomena.

However, one major disadvantage to generative models is their lack of control.
For unconditional generative models, in order to produce an output, one must (1) sample a noise vector and (2) supply this vector as input to a deep neural network.
If one would like to study outputs that contain certain features, then generation in this manner inhibits exploratory analysis, as it is nontrivial to predict what a given noise vector will map to under a generative model.
To make generative models more applicable and accessible to end-users, it is crucial to develop sampling methods that are not only more intuitive but also offer greater control over the sampling process, allowing users to focus on the features that are most relevant to their analysis.

As one concrete example, topology is a commonly studied feature in the visual analysis of vector fields.
Traditionally, topology is extracted from a provided field, and one studies the resulting topological representation for insight: a collection of critical points, and the relationships between these points.
But if one wants to understand what vector fields, within the distribution of the generative model, contain a particular topological structure, then it would normally be necessary to extensively draw samples from the generative model, and extract each field's topology for verification.
This limitation of generative models motivates the problem we study in this paper: given a topological description supplied by the user, we aim to generate a collection of vector fields that satisfy the topology.
By integrating domain-specific information, such as topology, into the generation process, we can create more informed and controllable generative models that align with the needs and expectations of end users, helping to facilitate the exploration and understanding of simulation ensembles.


We propose topology guidance, a method that augments the sampling process of a generative model, specifically a diffusion model~\cite{ho2020DenoisingDiffusion}, by incorporating topological information.
Such a feature-driven exploration of fields has precedence in visualization~\cite{takahashi2005feature}, in particular within the context of deep neural networks~\cite{han2022surfnet,cheng2018deep}.
However, a notable aspect of our approach is that we do not need to retrain, or fine-tune, an existing model to incorporate user-defined features.
We build on recent work of diffusion guidance~\cite{dhariwal2021DiffusionModels} to steer the denoising process of a diffusion model, done in such a manner that the output is aligned with the user's topology specification.
Our approach thus represents a training-free way to explore generative models based on topological features of interest, though in principle, our methodology can be extended beyond topology.

A basic problem addressed in our method is the following: how can we provide topology-based signals to a generative model?
Specifically, we aim to preserve user-specified critical points, varying by location, type (source, sink, saddle), and whether the critical point is stable or unstable.
To exert control over generation, we must consider how critical points are classified, and thus it is necessary to consider a vector field's spatial derivatives.
This implies that how we choose to represent a field is important: the representation should provide us with derivative information that can be used to guide generation.
To this end, our method builds on Functa~\cite{dupont2022DataFuncta}, wherein a coordinate-based neural network~\cite{sitzmann2020ImplicitNeural} and a denoising diffusion probabilistic model (DDPM) ~\cite{ho2020DenoisingDiffusion} are coupled together via a latent representation of fields.
The coordinate-based network we use, namely a field-conditioned SIREN~\cite{sitzmann2020ImplicitNeural}, is differential over both the spatial domain of the field, and the latent space in which fields are modeled.
We take advantage of these properties to derive an algorithm for diffusion guidance~\cite{dhariwal2021DiffusionModels}, where the iterative denoising of a latent vector is modified to satisfy properties of the field's spatial derivatives.
These properties are specifically designed to yield a particular critical point configuration, e.g. an unstable source.
Thus, by carefully balancing the topological signal from the SIREN and the denoising process of the diffusion model, our method is able to flexibly control the topological characteristics of the generated field, while ensuring the field lies within the modeled data distribution.
This combination allows for direct topological exploration of the generative model's output without requiring any modifications or retraining of the original model.

We evaluate our method on 2D vector fields that arise from an ensemble of fluid flows~\cite{Jakob2021}.
Experimentally we demonstrate how our method can successfully generate fields satisfying varying topological considerations, e.g. different kinds of critical points, arbitrarily positioned in the domain, as well as the specification of multiple critical points.
Importantly, these are not merely arbitrary fields that satisfy a user's topological description: we also show these fields remain close to the modeled data distribution of vector fields.
As an application of our method, we show how we may \emph{compare} distributions via topology guidance.
Namely, in considering two distributions of fluid flows that vary in Reynolds number, we show how topology guidance allows us to have a direct visual comparison of fields which have the same prescribed critical points, but otherwise vary in flow behavior,
as shown in Fig.~\ref{fig:teaser}.
We also show how topology guidance can allow one to study the distribution of fields \emph{limited} to a specific topological specification, and the topological complexity that results from their prescription.

In summary, we make the following contributions:
\begin{enumerate}
    \item We introduce topology guidance - a method for enhancing the sampling process of diffusion models by incorporating user-specified topology information, namely critical points. 
    \item We experimentally demonstrate the effectiveness of our method for a variety of topological descriptions.
    \item We demonstrate the benefits of our method for exploring a distribution conditioned on topology, as well as the use of topology guidance for comparing distributions.
\end{enumerate}


\section{Related work}

Our work primarily spans two research areas within visualization: (1) visualizing simulation ensembles, and (2) feature extraction for the visual analysis of fields. We discuss prior works in both areas, and how these methods relate to our approach.

\subsection{Visualizing simulation ensembles}

A simulation ensemble can produce a massive amount of data, often taking the form of sampled fields that vary by spatial position, time, and a parameter space underlying the problem of study~\cite{wang2018visualization}, e.g. for a fluid flow this might be viscosity.
Given the high computation cost of simulation, coupled with the large size of the data, methods for visualizing ensembles typically focus on aspects of data reduction~\cite{li2018data}.
This has been a key motivation in applying deep learning techniques to meet scientific visualization needs, please see Wang et al.~\cite{wang2022DL4SciVisStateoftheArt} for a detailed survey.
Specifically, given a reduced representation of the ensemble, the goal is to synthesize novel fields that serve as good approximations to the actual simulated data.
Methods for superresolution are commonly used to address this problem, wherein the main goal is to synthesize higher-resolution detail across space and/or time~\cite{Jakob2021,han2022STNetEndtoEnd,shen2024PSRFlowProbabilistic}.
Often these models are trained on fields sampled at a high-resolution, those corresponding to simulations sampled over a parameter space, while the provided input corresponds to data subsampled in space/time for novel simulation parameters.

Prior methods for superresolution are further distinguished by the data representation of the output.
Specifically, gridded data tends to rely on CNN-based superresolution~\cite{wurster2022deep,han2022STNetEndtoEnd}.
In contrast, coordinate-based neural representations of fields, commonly known as implicit neural representations (INR)~\cite{tancik2020FourierFeatures,sitzmann2020ImplicitNeural}, allows for the evaluation of a field at arbitrary inputs within the domain.
In particular, INRs have recently been shown to be beneficial for superresolution in numerous contexts~\cite{han2023CoordNetData,tang2024STSRINRSpatiotemporal}, e.g. training an INR on subsampled data, and evaluating on arbitrary locations as a form of superresolution.
Our approach also relies on INRs as the data representation, in part for their compact representation of fields, but more importantly they offer differentiable representations of fields~\cite{sitzmann2020ImplicitNeural}.
We take advantage of this in the guided generation of fields.

Although superresolution methods are effective for addressing problems of data size in simulation ensembles, the requirement of subsampled data often assumes a full simulation has been run, and consequently down-sampled, thus still incurring high computation cost.
Thus, numerous visualization methods exist for building \emph{surrogate} models of simulations, typically designed to allow for interactive visual analysis.
These methods can be viewed as interpolation schemes: given parameters for a simulation as input, the goal is to output a field that would have been, equivalently, produced by a simulation for the specified parameters.
Approaches range in modeling the space of visualization outputs~\cite{he2019InSituNetDeep}, view-dependent prediction of fields~\cite{shi2022vdl}, as well as attempting to model the full parameter space~\cite{shi2022gnn} of a simulation.

Surrogate models make an assumption that the field output by a simulation is uniquely identified by a small set of parameters.
However, the modeling of certain phenomena, for instance models of turbulence, often assume initial conditions that are random in nature, e.g. noise of prescribed frequency for turbulent fluid flows~\cite{Jakob2021}.
In such scenarios, surrogate models must account for the inherent randomness, and deep generative models have emerged as an appealing choice.
The generative modeling of a distribution of fields has witnessed significant recent work within scientific computing, ranging from domains such as material science~\cite{hsu2021microstructure,lee2023microstructure}, turbulence~\cite{drygala2022generative,lienen2024zero} and combustion~\cite{laubscher2020application,shin2023probabilistic}.
Thus, in contrast with simulation-based exploration methods~\cite{he2019InSituNetDeep,shi2022gnn} that assume a \emph{deterministic} mapping from parameters to fields, generative models produce \emph{random} instances from a modeled data distribution.
Topology guidance aims to better support the visual analysis of such generative models, making generation more controllable.

\subsection{Features in vector fields}

Our method is related to several types of approaches in utilizing features for vector fields: the extraction of features, querying fields based on features, as well as feature-driven design of vector fields.

Feature extraction in vector fields has witnessed significant development in the visualization community, please see the following surveys~\cite{laramee2007topology,pobitzer2011state} for an overview.
Features can be largely distinguished by their representation, e.g. individual points~\cite{weinkauf2005extracting}, curves~\cite{hassouna2007extraction} that serve to connect feature points, or areas~\cite{gunther2018state} that serve to partition the domain into homogeneous regions of flow.
Topology-based feature extraction is concerned with extracting critical points in the domain, namely those locations where the vector's components are zero.
A critical point can further be categorized by the vector field's behavior in a local region, namely by computing the Jacobian of the vector field and inspecting its eigenvalues, and determinant~\cite{HelmanVisualVectorTopo1991}.
A vector field's collection of critical points can serve as a visual summary of the field, indicating regions where flows originate, converge, as well as identify regions of swirling motion.
Topology is not the only way to derive point-based features; vortex extraction~\cite{deng2019cnn} is another common approach for feature extraction.
As in critical points, the criteria of a vortex often requires access to the Jacobian of a vector field.
This commonality in using vector field derivatives to extract features is one of the main motivations in our work, as we wish to study how such information can be utilized to guide generative models of vector fields.

Topology guidance can be seen as a method for \emph{querying} fields based on user-specified features.
Numerous methods exist to query steady, or unsteady, flow fields, often focused on detecting regions that contain prescribed features.
For instance, the specification of exemplar streamlines~\cite{theisel2005topological}, pathlines~\cite{theisel2005topological,hong2014flda}, or quantities extracted within a spatiotemporal region~\cite{dallmann1995flow} can all serve as queries in finding regions of interest to the user.
Within the context of an ensemble simulation, these query-based methods can be suitable in helping discover features within a distribution of fields.
However, there are two main limitations with such an approach.
First, we require access to the original vector fields, which can be costly for large-scale datasets - indeed, generative models can be viewed as a form of dataset compression~\cite{santurkar2018generative,kingma2021variational}.
Secondly, limited to the set of fields sampled from a distribution, what a user specifies as a query might not exist in the given data, e.g. a critical point, at a specific location of a particular type.
On the other hand, such a query might very well exist within the data distribution of a generative model data, and thus more controllable generation schemes, guided by user-defined features, can help address this problem.

The approach taken by topology guidance is closely related to methods for designing fields, e.g. whether vector fields~\cite{theisel2002designing,chenDesign2DTimeVarying2012} or tensor fields~\cite{zhangInteractiveTensorField2007}.
Such approaches are often topology-driven, wherein a user specifies a particular topological configuration as input, and the output is designed to be a field that satisfies a user's specified topology.
However, these methods are not suitable for understanding a distribution of vector fields, as they are defined independently of a data distribution.
For instance, when provided a specific topology as input, two distributions of vector fields might give quite different fields that, nevertheless, satisfy the given topology.
These types of scenarios motivate our work, namely we aim to synthesize vector fields that not only satisfy a user's specified topology, but also ensure the generated fields are within a modeled data distribution.

\section{Background}
\label{sec:background}
In this section, we introduce the motivation behind topology guidance and discuss prior works on which we base our method.
Our approach is motivated by datasets commonly produced in modern numerical simulations, namely, fields that vary over a parameter space of interest to a domain scientist.
As the cost of running and storing simulations can be quite expensive, recent work has sought to build surrogate models~\cite{he2019InSituNetDeep,shi2022gnn} that aim to produce fields, or visualizations therein, through conditioning on simulation parameters as input to a deep neural network.
This is a feasible approach when parameters are low-dimensional, and carry intuition for the domain scientist, e.g. viscosity of a fluid.
On the other hand, a simulation might contain high-dimensional parameters that are randomly set by domain experts, e.g. the initial state of a time-varying system.
In these cases, randomly-assigned parameters cannot be easily treated as deterministic input to a neural network.
Thus, numerous methods recently have sought to build generative models over the possible outcomes of a simulation~\cite{drygala2022generative,lienen2024zero}, leading to a compact, yet faithful, representation of a simulation for domain scientists to study.



However, the utility of a generative model for post-hoc analysis faces a fundamental limitation: due to the randomness inherent in sampling from a generative models, end users have little control on the generation of fields.
Put simply, a user does not know what they will see when they draw a sample from a generative model.
Given this limitation, it is natural to ask: what types of controls make sense to provide users when interacting with generative models?
In this paper, we argue that the types of features used to normally \emph{study} fields should be purposed as inputs in the \emph{generation} of fields.
We consider this within the context of topological features of vector fields, namely critical points.
The ability to synthesize a vector field of a prescribed topology, varying by location, type, and the number of critical points, has numerous benefits.
First, we can better facilitate the \emph{comparison} of two distributions, each represented as generative models.
The ability to specify topology that fields, across distributions, have in common helps with issues of \emph{spatial alignment} between generated fields.
Namely, we can study the local behavior around their common topological features, and identify similarities/differences between fields, e.g. varying complexity due to differences in parameters.
Secondly, we may perform more fine-grained \emph{what-if} analyses on existing fields, using the topology extracted from a given field as its characteristic ``shape'', and asking how this field would look under a different data distribution.
For instance, we may want to know whether an extracted topology (e.g. a point of repelling focus) induces more/less turbulence under a different distribution.

One potential approach to achieve such control is to sample the ensemble simulation data and extract topological features, 
with the aim of collecting a large set of fields that satisfy one's prescribed features.
However, this type of analysis is limited by the available samples, and even if a large collection of samples is generated, there is no guarantee that the feature of interest will be present within this collection.
To overcome these challenges, we propose foregoing such trial-and-error sampling, and instead incorporate topology-based features within the sampling process itself.
To this end, we summarize three key requirements for extracting a field with certain topological features: 
\begin{itemize}
    \item (\textbf{R1}) The generated field must satisfy the desired topological features, e.g., the presence of specific critical points (sources, sinks, and saddles) and their behavior (e.g. stable vs. unstable).
    \item (\textbf{R2}) The generated field should be within the distribution of the simulation, ensuring that the characteristics of the simulation data are preserved in the output.
    \item (\textbf{R3}) We should not require additional training of the generative model to incorporate ones desired topological features.
\end{itemize}

To meet the three design requirements outlined above, our approach leverages the following:
\textit{coordinate-based neural networks}, \textit{diffusion models}, and \textit{diffusion guidance}.
In the following subsections, we provide background on these areas, highlighting how they can help us realize the objective of topology guidance.

\subsection{Coordinate-based neural network}
\label{eq:coord}

Vector field topology is usually defined by particular constraints placed on the spatial derivatives of a vector field.
Thus, how we decide to represent a field in our generative model is important: the representation should provide us a way to access, and utilize, spatial derivatives in the pursuit of satisfying a prescribed topology specification.
To this end, we build off of the approach of Functa~\cite{dupont2022DataFuncta}, a coordinate-based neural network that can generalize to fields by conditioning on field-specific latent vectors.
Functa builds off of SIREN~\cite{sitzmann2020ImplicitNeural}, a coordinate-based network that has shown effective for derivative-based processing, e.g. solving the Poisson and Helmholtz equations.
Indeed, forming the gradient of a SIREN results in another SIREN, and thus is suitable in providing a topology-based signal for guiding generation (\textbf{R1}).

In detail, Functa is a multi-layer perceptron, wherein the first layer yields a sinusoidal positional encoding of a given element $\mathbf{p} \in \mathbb{R}^2$ in the domain,
\begin{equation}
    \mathbf{x}^{(0)} = \sin (\mathbf{W}^{(0)} \mathbf{p} ),
\end{equation}
where $\mathbf{W}^{(0)} \in \mathbb{R}^{D' \times 2}$ is a learnable weight matrix, and all subsequent layers condition on a field-specific latent code:
\begin{equation}
    \mathbf{x}^{(l)} = \sin (\mathbf{W}^{(l)} \mathbf{p} + \mathbf{M}^{(l)} \mathbf{z}).
\end{equation}
A field is represented by a latent vector, denoted $\mathbf{z} \in \mathbb{R}^D$, subsequently mapped to the space of a SIREN via a learned linear transformation $\mathbf{M}^{(l)} \in \mathbb{R}^{D' \times D}$,
as shown in Fig. ~\ref{fig:background} (B).
The vector $\mathbf{w}^{(l)}$ serves to modulate the SIREN such that it may generalize over both (1) positions in the domain ($\mathbf{p}$), and (2) fields ($\mathbf{z}$) that span a distribution.
In summary, our representation of vector fields takes the form of a field-conditional INR $f_{\phi} : \mathbb{R}^2 \times \mathbb{R}^D \rightarrow \mathbb{R}^2$, taking as input (1) a position $\mathbf{p}$ in the domain, (2) a latent vector $\mathbf{z} \in \mathbb{R}^D$ that represents a field, and outputs a 2D vector, with $\phi$ containing all learnable parameters $\mathbf{W}^{(l)}$ and $\mathbf{M}^{(l)}$

Beyond generalizing to arbitrary fields, Functa has two key features:
\begin{enumerate}
    \item Functa can rapidly find the latent vector representation of each field via model-agnostic meta-learning~\cite{finn2017model};
    \item Functa can learn a meaningful latent space by extracting shared features across multiple fields.
\end{enumerate}
The first point ensures that, upon conclusion of meta-learning, we can quickly find latent vectors for fields drawn from a distribution, e.g. a simulation ensemble.
More importantly, the second point ensures that latent vectors found via gradient descent are well-structured, e.g. two fields that are similar will contain similar latent representations.
Functa~\cite{dupont2022DataFuncta} takes advantage of this for building generative models over latent spaces, which we next discuss.

\subsection{Diffusion models}

\begin{figure}[!t]
    \centering
    \includegraphics[width=1\linewidth]{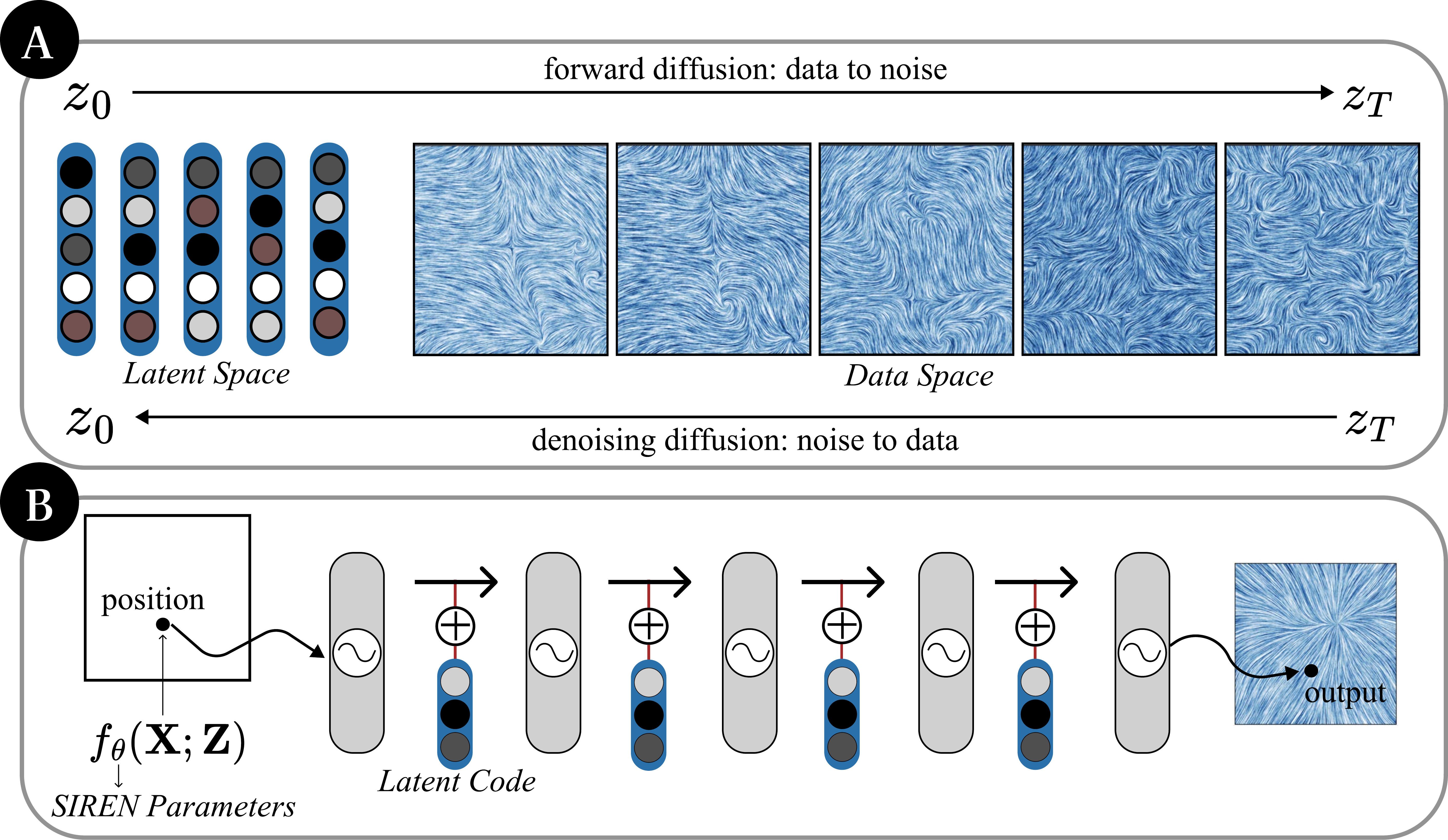}
    \caption{
    We present an overview of the model we use for topology guidance. (A) A diffusion model operates in a compact latent space, performing forward and denoising diffusion. (B) Mapping sampled latent vectors back to the data space using a pre-trained SIREN. The latent vector modulates the SIREN, which takes an arbitrary position in the flow field as input and outputs the vector field value at that point.
    }
    \label{fig:background}
\end{figure}

For our method, we assume as input a set of fields that have been gathered from a simulation ensemble.
Given that we only have a partial view of the full space of fields, we aim to build a model that can synthesize novel vector fields, ideally those that can equivalently be obtained via simulation.
To this end, we build on Functa and build a generative model, namely a denoising diffusion probabilistic model~\cite{ho2020DenoisingDiffusion}, over the field-based latent representations $\mathbf{z}_n \in \mathbb{R}^D$.
Indeed, diffusion models have been adapted to a number of domains in computational science~\cite{buehler2023predicting,jing2023eigenfold,lienen2024zero}, and in particular, diffusion models of latent data representations have shown quite effective as scalable, yet effective, generative models~\cite{rombach2022high} (\textbf{R2}).

In detail, given a vector drawn from a fixed distribution, usually a standard normal distribution $\mathbf{z}_T \sim \mathcal{N}(\mathbf{0}, \mathbf{I})$, a diffusion model performs iterative denoising until we arrive at a vector that is intended to belong to the data distribution, e.g. in our case $\mathbf{z}_0$ would correspond to a plausible vector field in the data distribution of a simulation.
As shown in Fig. ~\ref{fig:background} (A),
a diffusion model is defined by a fixed forward process which iteratively adds noise to a given vector from the dataset $\mathbf{z}_0$:
\begin{equation}
    \mathbf{z}_t = \sqrt{\bar{\alpha}_t} \mathbf{z}_0 + \sqrt{1-\bar{\alpha}_t} \be \; , \; \be \sim \mathcal{N}(\mathbf{0}, \mathbf{I}),
    \label{eq:forward}
\end{equation}
where $\mathbf{z}_t$ represents the noisy sample at step $t$, $\bar{\alpha}_t$ is a noise schedule parameter, and $\be$ is a vector drawn from a standard Gaussian.
A reverse diffusion process seeks to learn a denoising model $\be_{\theta}$ comprised of parameters $\theta$, wherein given a vector drawn from a standard normal $\be \sim \mathcal{N}(\mathbf{0}, \mathbf{I})$, a diffusion model seeks to best approximate this noise vector:
\begin{equation}
\be_{\theta}(\mathbf{z}_t, t) \approx \be = \frac{\mathbf{z}_t - \sqrt{\bar{\alpha}_t}}{\sqrt{1-\bar{\alpha}_t}}.
\end{equation}

Once a diffusion model is trained, there is considerable flexibility in how to draw samples from the model.
What will be beneficial for topology-based guidance is the method of denoising diffusion implicit models (DDIM)~\cite{song2022DenoisingDiffusion}.
Specifically, the forward diffusion process (c.f. Eq.~\ref{eq:forward}) induces a distribution on the reverse process, denoted $q(\mathbf{z}_{t-1} | \mathbf{z}_t , \mathbf{z}_0)$, e.g. to obtain a denoised latent vector $\mathbf{z}_t$, we must condition on (1) the previous, slightly noisier latent $\mathbf{z}_t$, and the original latent vector $\mathbf{z}_0$.
For generation, we do not have access to $\mathbf{z}_0$, but instead we can approximate it via Eq.~\ref{eq:forward}:
\begin{equation}
    \mathbf{\hat{z}}_t = \frac{\mathbf{z}_t - \baralphaone \be_{\theta}(\mathbf{z}_t, t)}{\baralpha}.
    \label{eq:reverse}
\end{equation}
This prediction is then used in place of $\mathbf{z}_0$ in the reverse diffusion process $q(\mathbf{z}_{t-1} | \mathbf{z}_t , \mathbf{\hat{z}}_t)$, in order to obtain the next sample $\mathbf{z}_{t-1}$.

\subsection{Diffusion guidance}
\label{subsec:guidance}

Our work builds off of the recent research in controlling the sampling process of diffusion models~\cite{dhariwal2021DiffusionModels,hoClassifierFreeDiffusion}.
Specifically, it is possible to draw samples from a diffusion model that satisfy certain user-prescribed properties (\textbf{R3}).
This can be achieved in a training-free manner by revising the standard denoising process (c.f. Eq.~\ref{eq:reverse}).
This allows for considerable control over what a user can generate in a diffusion model, and has been studied extensively within computer vision for classifier-based guidance~\cite{dhariwal2021DiffusionModels,hoClassifierFreeDiffusion}, text-based guidance~\cite{rombach2022high,bansal2023UniversalGuidance}, and more broadly the shape, location, and appearance of output images~\cite{epsteinDiffusionSelfGuidance}.

The basic principle underlying diffusion guidance is the connection between denoising diffusion models and denoising score matching~\cite{vincent2011connection}.
Specifically, the predicted noise in a diffusion model $\be_{\theta}(\mathbf{z}_t, t)$ serves as an approximation to the negative score function for the noised data distribution, or the gradient of the log likelihood of the data distribution.
This implies that we can take the predicted noise, and combine it with a separate term that reflects a property of the data we would like to preserve in the generated output, e.g. not just where the data density is high.
We denote this as a guidance function, $\mathcal{E}(\mathbf{z}_t, t ; C)$ which considers the current noise vector in the diffusion process ($\mathbf{z}_t$), and user-supplied constraints about the output contained in set $C$.
Assuming this guidance function is differentiable, then we can modify the denoising process via descending on its gradient:
\begin{equation}
    \tilde{\be}(\mathbf{z}_t, t) = \be(\mathbf{z}_t, t) + \omega \nabla_{\mathbf{z}_t} \mathcal{E}(\mathbf{z}_t, t ; C),
    \label{eq:guidance}
\end{equation}
where $\omega$ is a hyperparameter that represents the guidance strength, e.g. the extent to which denoising is influenced by guidance.

Our method of topology guidance builds on this formulation, wherein the set $C$ represents a topology specification from a user.
However, for topology-based guidance to work, we must design an energy function that gives us some kind of signal about the topological information of interest to the user.
Towards this end, key to our method is the use of a coordinate-based network (c.f. Sec.~\ref{eq:coord}), which bridges (1) topology-based signals provided by a SIREN with (2) guidance of the generated field.



\section{Topology guidance}

In this section we present our approach for topology guidance, namely, how to guide a diffusion model of vector fields towards a user-specified topology configuration. We organize topology guidance along the following factors:
\begin{enumerate}
    \item \textbf{Critical point}: ensuring that a critical point exists at a specified location.
    \item \textbf{Type of critical point}: distinguishing a critical point whether it is attracting, repelling, or a saddle.
    \item \textbf{Stability of a critical point}: attracting \& repelling critical points can be further categorized as stable, or unstable.
\end{enumerate}
A key aspect of our approach is the \emph{composition} of the above factors, e.g. a user may wish to study a distribution of fields that merely have some type of critical point at a given location; alternatively, they may be highly precise in their specification.
We discuss each of these factors step by step, namely, how to derive suitable guidance functions (c.f. Sec.~\ref{subsec:guidance}).

\subsection{Prescribing a critical point}

The most general form of guidance is simply specifying that a critical point exists at a given location.
A critical point is defined as a location in the domain for which the vector's components are all zero.
This can equivalently be phrased as the vector's norm being zero at the location.
Thus, we can derive a suitable guidance function by prescribing that the norm of the vector field should be minimized at a given location:
\begin{equation}
    \mathcal{E}_c(\mathbf{z}_t , t; C) = \lVert f_{\phi}(\mathbf{p} ; \mathbf{z}_t) \rVert,
    \label{eq:prescribe}
\end{equation}
where the set $C = (\mathbf{p})$ simply contains the location $\mathbf{p} \in \mathbb{R}^2$ in the domain where the user intends to introduce a critical point.

The guidance function $\mathcal{E}_c$ is used to modify the noise normally predicted by a diffusion model (c.f. Eq.~\ref{eq:guidance}).
In practice, this has a \emph{localized} effect on the output vector field, please see Fig.~\ref{eq:prescribe} for an illustration.
Specifically, fixing the initial noise vector $\mathbf{z}_T$, guidance has the effect of changing the vector field in a local vicinity of the prescribed location, e.g. to impart a critical point.
Thus, for the region of space \emph{away} from the critical point, the overall behavior of the vector field will remain similar to the vector field that would have been produced \emph{without} guidance.
This ensures the generated vector field remains within the modeled data distribution.

A complication that arises in the above is that the latent vector supplied to the SIREN in Eq.~\ref{eq:prescribe} corresponds to a \emph{noisy} latent vector, where the level noise depends on the time step $t$ in the reverse diffusion process.
When $t$ is large, we find that the supplied vector $\mathbf{z}_t$ is out-of-distribution for the SIREN, and thus cannot give a meaningful signal on the critical point.

We address this problem in two ways.
First, following Bansal et al.~\cite{bansal2023UniversalGuidance}, we build off of the DDIM sampling approach in predicting a clean latent vector $\mathbf{\hat{z}}_t$ given its noisy counterpart $\mathbf{z}_t$.
The vector $\mathbf{\hat{z}}_t$ is much more likely be recognized as in-distribution for the SIREN, even though at an intermediate time step $t$ it might not yet correspond to a vector field within the data distribution.
Nevertheless, we may use this in place of $\mathbf{z}_t$ in computing the guidance function:
\begin{equation}
    \mathcal{E}_c(\mathbf{z}_t , t; C) = \lVert f_{\phi}(\mathbf{p} ; \mathbf{\hat{z}}_t) \rVert.
    \label{eq:prescribe}
\end{equation}
We note that guidance still relies on gradients with respect to the noisy vector, e.g. $\nabla_{\mathbf{z}_t} \lVert f_{\phi}(\mathbf{p} ; \mathbf{\hat{z}}_t) \rVert$, in order to properly arrive at the next denoised version $\mathbf{z}_{t-1}$.

\begin{figure}[!t]
    \centering
    \includegraphics[width=1\linewidth]{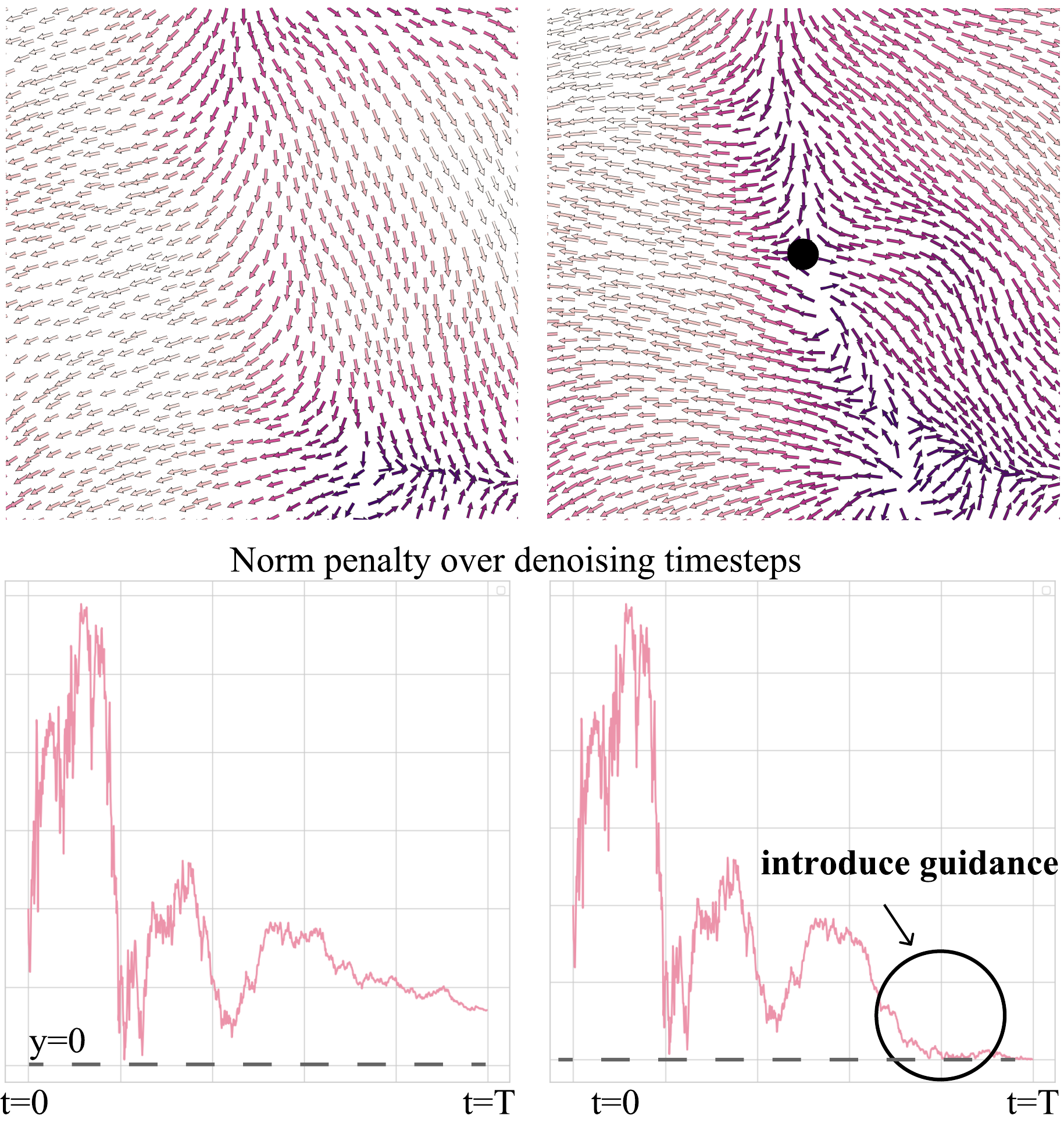}
    \caption{
    Localized effect of guidance on vector field generation.
Left: Sample without guidance. Right: Sample with guidance. The line graph shows the norm change at the user-specified location over denoising timesteps. Guidance is injected at time step 600. After prescribing a critical point through guidance, the norm at the specified location significantly decreases to near 0.
    }
    \label{fig:methods-norm}
\end{figure}

The second way in which we address this problem involves the choice of time steps at which to introduce guidance.
Since the denoising process starts from random noise, the initial steps of denoising may produce highly distorted samples, where the predicted clean latent vector $\mathbf{\hat{z}}_0$ is still of little use.
Therefore, it is not desirable to introduce guidance too early in the process. 
On the other hand, introducing guidance too late may be ineffective because the local behavior of the sample has already formed and becomes difficult to modify. 
Through experimentation, we have carefully determined the time step at which to introduce guidance, based on the observation that the vector norm starts to decrease and gradually approaches zero at a certain point during the denoising process; this is highlighted in ~\cref{fig:methods-norm}.
We summarize the sampling process incorporated with our guidance function in ~\cref{alg:sample}.

\begin{algorithm}[H]
\caption{Topology guidance for prescribing a critical point}
\label{alg:sample}
\begin{algorithmic}[1]
\Statex \textbf{Input}: Prescribed location of critical point $\mathbf{p} \in \mathbb{R}^2$
\Statex \textbf{Output}: Vector field with specified topology \textbf{$z$}
\State $\mathbf{z}_T \sim \mathcal{N}(\mathbf{0}, \mathbf{I})$
\For{$t = T, ..., 1$} 
    \State $\be \sim \mathcal{N}(\mathbf{0}, \mathbf{I})$ if $t > 1$, else $\be = \mathbf{0}$ 
    \State $\tilde{\be}_{\theta}(\mathbf{z}_t, t) = \be_{\theta}(\mathbf{z}_t, t) + \omega \nabla_{\mathbf{z}_t} \lVert f_{\phi}(\mathbf{p} ; \mathbf{\hat{z}}_t) \rVert $
    \State $\mathbf{z}_{t-1} = \frac{1}{\sqrt{\bar{\alpha}_t}} \left( \mathbf{z}_t - \frac{1-\bar{\alpha}_t}{\sqrt{1-\bar{\alpha}_t}} \tilde{\epsilon}_{\theta}(\mathbf{z}_t, t) \right) + \sigma_t \be$ 
\EndFor
\State \Return $\mathbf{z}_0$
\end{algorithmic}
\end{algorithm}

\begin{figure}[!t]
    \centering
    \includegraphics[width=1\linewidth]{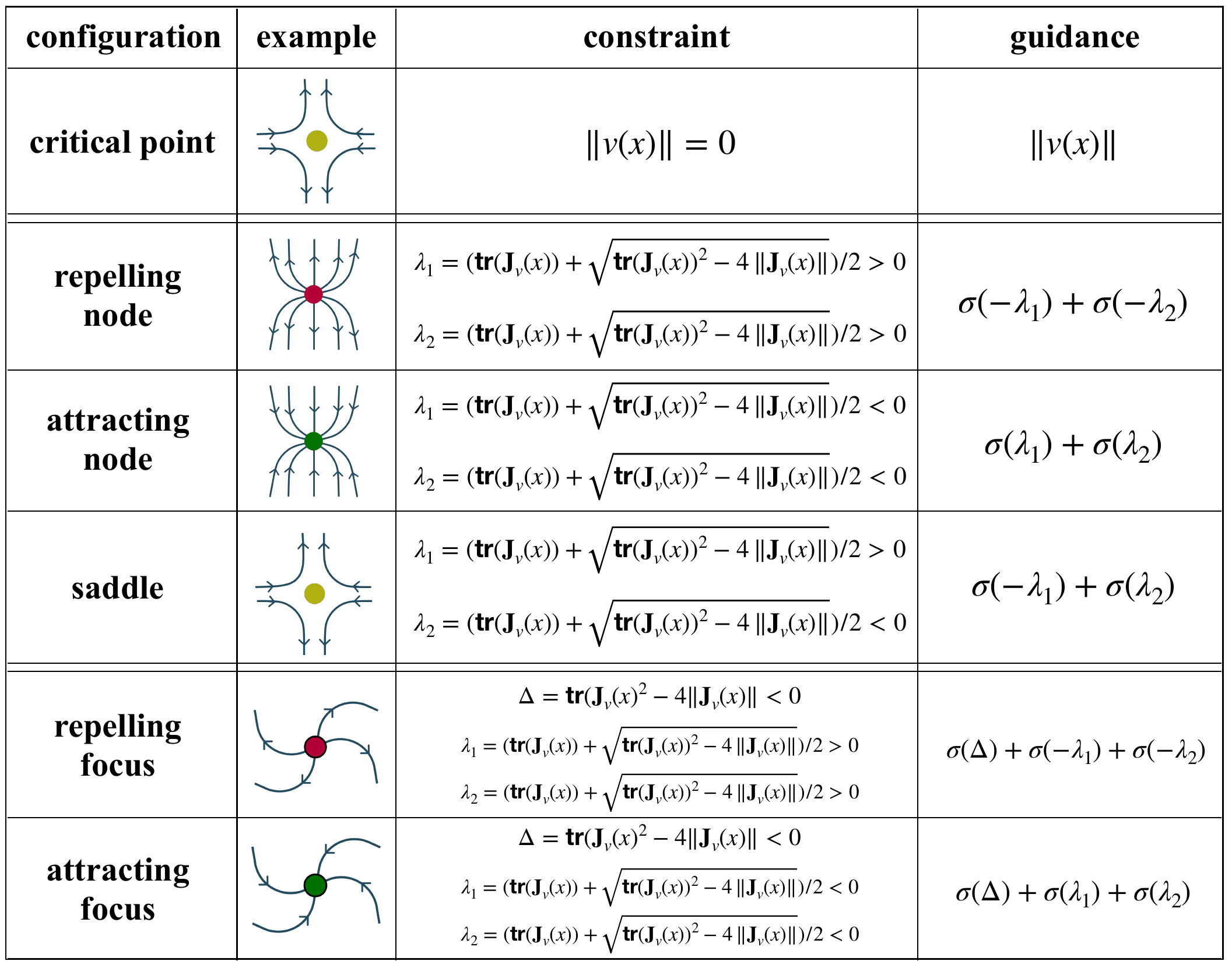}
    \caption{
    Our method provides a general framework for drawing vector fields from a generative model containing a variety of topology specifications, namely critical points. The figure summarizes five configurations of vector field critical points\cite{HelmanVisualVectorTopo1991}, along with examples, necessary constraints, and corresponding energy functions in our approach.
    }
    \label{fig:methods-guidance}
\end{figure}

\begin{figure}[!t]
    \centering
    \includegraphics[width=1\linewidth]{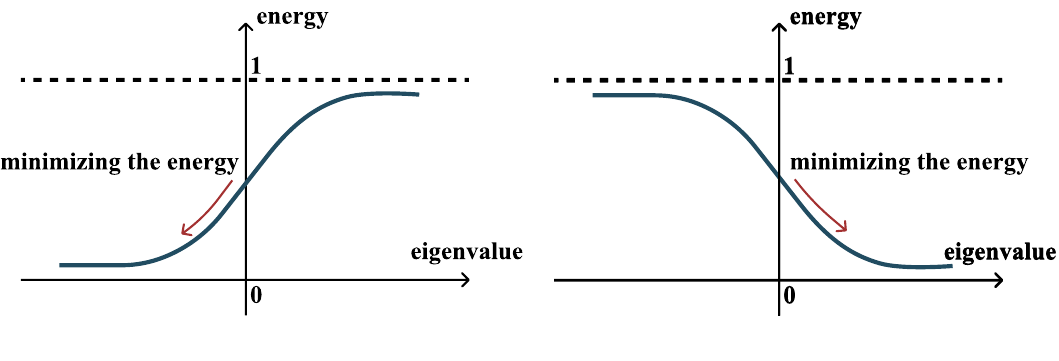}
    \caption{
Manipulating critical point types using an energy function inspired by the sigmoid function. By optimizing the energy function to control the eigenvalues signs of the Jacobian matrix at specified locations, a guide can generate a sink (left) or a source (right).
    }
    \label{fig:methods-energy}
\end{figure}

\subsection{Prescribing critical point type}

We build on the specification of a critical point, allowing for a more fine-grained specification along the \emph{type} of critical point.
Critical points can be classified according to the local behavior of flow, namely whether a point is attracting or repelling, or serving as a saddle.
Categorizing a critical point into these three types requires access to the Jacobian of the vector field:
\begin{equation}
J_{\phi}(\mathbf{p} ; \mathbf{\hat{z}}_t) =
\begin{bmatrix}
\frac{\partial f_{\phi}(\mathbf{p} ; \mathbf{\hat{z}_t})_x}{\partial x} & \frac{\partial f_{\phi}(\mathbf{p} ; \mathbf{\hat{z}_t})_x}{\partial y} \\
\frac{\partial f_{\phi}(\mathbf{p} ; \mathbf{\hat{z}_t})_y}{\partial x} & \frac{\partial f_{\phi}(\mathbf{p} ; \mathbf{\hat{z}_t})_y}{\partial y}
\end{bmatrix}.
\end{equation}
Categorization follows by extracting the eigenvalues $\lambda^{(1)}_{\phi}(\mathbf{p} ; \mathbf{\hat{z}}_t)$ and $\lambda^{(2)}_{\phi}(\mathbf{p} ; \mathbf{\hat{z}}_t)$ of the Jacobian, and classifying based on the signs of their real-valued components:
\begin{equation}
J_{\phi}(\mathbf{p} ; \mathbf{\hat{z}}_t) :
\begin{cases}
\text{Sink},  & \lambda^{(1)}_{\phi}(\mathbf{p} ; \mathbf{\hat{z}}_t) < \lambda^{(2)}_{\phi}(\mathbf{p} ; \mathbf{\hat{z}}_t) < 0\\
\text{Source}, & 0 < \lambda^{(1)}_{\phi}(\mathbf{p} ; \mathbf{\hat{z}}_t) < \lambda^{(2)}_{\phi}(\mathbf{p} ; \mathbf{\hat{z}}_t) \\
\text{Saddle}, & \lambda^{(1)}_{\phi}(\mathbf{p} ; \mathbf{\hat{z}}_t) < 0 < \lambda^{(2)}_{\phi}(\mathbf{p} ; \mathbf{\hat{z}}_t)
\end{cases}    
\end{equation}
We make explicit the dependency on the parameters $\phi$: eigenvalues can be computed in closed-form given the differentiability of the SIREN.
We further note the dependency on the predicted latent code $\mathbf{\hat{z}}_t$, and thus the eigenvalues at a given location evolve over the course of denoising.

A guidance function for critical point type should serve to push the signs of the eigenvalues towards what is necessary for the prescribed type.
To this end, we treat this as a classification problem, wherein we apply the sigmoid function to the eigenvalues.
This has an interpretation of treating eigenvalues as logits, and prescribing the signs of logits via minimization of the sigmoid.
Specifically, our guidance function can be expressed as:
\begin{equation}
    \mathcal{E}_t(\mathbf{z}_t , t; C) = \sigma\left(\beta_1 \cdot \lambda^{(1)}_{\phi}(\mathbf{p} ; \mathbf{\hat{z}}_t) \right) + \sigma\left(\beta_2 \cdot \lambda^{(2)}_{\phi}(\mathbf{p} ; \mathbf{\hat{z}}_t) \right)
    \label{eq:type},
\end{equation}
where the set $C = (\mathbf{p}, \beta_1, \beta_2)$ contains the location of the critical point, but in addition, information on critical point types.
Specifically, a repelling point will be achieved with $\beta_1 = \beta_2 = -1$, an attracting point with $\beta_1 = \beta_2 = 1$, and a saddle with $\beta_1 = -1 , \beta_2 = 1$.
Over the denoising process, the sum of the sigmoids will be continually minimized, driving the logits towards their appropriate signs, as illustrated in ~\cref{fig:methods-energy}.


\subsection{Prescribing critical point stability}

In addition to specifying the type of critical point, we can also control for the stability of a critical point, distinguished by stable, and unstable, critical points.
Stability can be categorized according to the Jacobian matrix, requiring the computation of its trace, and determinant.
Specifically, we may categorize stability based on the following quantity:
\begin{equation}
    \Delta_{\phi}(\mathbf{p} ; \mathbf{\hat{z}}_t) = \Tr(J_{\phi}(\mathbf{p} ; \mathbf{\hat{z}}_t))^2 - 4 | J_{\phi}(\mathbf{p} ; \mathbf{\hat{z}}_t) |.
\end{equation}
The sign of $\Delta$ determines the type of stability: if positive, then we have a stable critical point; otherwise we have negative critical point.

We design a guidance function for the property of stability in the same manner as the type.
Specifically, we have:
\begin{equation}
    \mathcal{E}_s(\mathbf{z}_t , t; C) = \sigma\left(\beta_s \cdot \lambda^{(1)}_{\phi}(\mathbf{p} ; \mathbf{\hat{z}}_t) \right)
    \label{eq:stability},
\end{equation}
with set $C = (\mathbf{p}, \beta_s)$ now containing information on stability.
Specifically, an unstable critical point will be achieved via $\beta_s = 1$, and a stable critical point with $\beta_s = -1$.

\subsection{Combining guidance functions}

\begin{figure}[!t]
    \centering
    \includegraphics[width=1\linewidth]{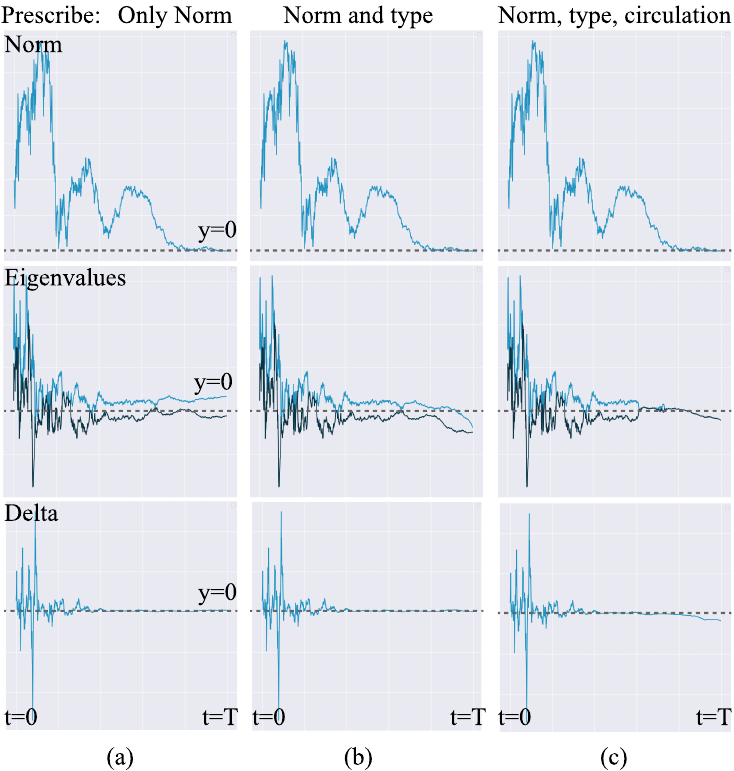}
    \caption{
    An energy function that applied to norms, eigenvalues, and $\Delta$ can work independently. Moreover, it is also possible to combine these components to create a more comprehensive energy function.
    }
    \label{fig:methods-combination}
\end{figure}

\textbf{Combination of Guidance}. 
To achieve finer control over the generated vector fields, we can combine the specification of (1) critical point type and (2) stability.
This is accomplished by composing~\cref{eq:prescribe,eq:type,eq:stability}, realized by adding guidance functions together:
\begin{equation}
\begin{aligned}
    \mathcal{E}_{com}(\mathbf{z}_t , t; C) = & \lVert f_{\phi}(\mathbf{p} ; \mathbf{\hat{z}}_t) \rVert \\
    & + \sigma\left(\beta_1 \cdot \lambda^{(1)}_{\phi}(\mathbf{p} ; \mathbf{\hat{z}}_t) \right) + \sigma\left(\beta_2 \cdot \lambda^{(2)}_{\phi}(\mathbf{p} ; \mathbf{\hat{z}}_t) \right) \\
    & + \sigma\left(\beta_s \cdot \lambda^{(1)}_{\phi}(\mathbf{p} ; \mathbf{\hat{z}}_t) \right)
        \label{eq:combination}
\end{aligned}        
\end{equation}
This linear combination of guidance functions is effective because it allows for a balanced influence of both type and stability constraints in the denoising process. 
As shown in \cref{fig:methods-combination}, the combined guidance function can control the norm, eigenvalues of the Jacobian matrix, and $\Delta$, 
satisfying the constraints without introducing conflicting interactions between them.
This flexibility enables us to generate a wide range of vector field patterns with more detailed characteristics.


\textbf{Multiple Critical Points}.
It is, further, straightforward to specify multiple critical points throughout the domain:
\begin{equation}
\mathcal{E}(\mathbf{z}_t , t; \mathcal{C}) = \sum_{C \in \mathcal{C}} \mathcal{E}_{com}(\mathbf{z}_t , t ; C),
\end{equation}
where $\mathcal{C}$ is a set consisting of critical point specifications, namely, any potential combination previously discussed.

\section{Evaluation}
\label{sec:evaluation}
In this section, we introduce the evaluation of topology guidance. The experiment setup provides details on the datasets, implementation, and parameters, enabling reproducibility. Quantitative evaluation employs metrics to demonstrate the effectiveness of our approach, while qualitative results visually illustrate its applicability. The combination of quantitative and qualitative evidence supports the effectiveness and intuition of topology guidance.
\subsection{Experiment Setup}

\textbf{Dataset}. Our dataset is from \cite{Jakob2021}.
The dataset is a time-varying 2D vector field with a resolution of $512\times512\times1024$. 
Our focus is on the critical points within its topological structure, which is a time-independent local feature. 
We first detect the topology feature density of fields with different sample resolutions. 
It indicates that for fields with a sample resolution of $256\times 256$, when randomly sampling 1000 fields, and most of them possess at least one fixed point.
Therefore, we use the crop of resolution $256\times 256$ to serve as the training dataset.

\textbf{Modulated INR training details.}
We follow the three training suggestions proposed by Functa\cite{dupont2022DataFuncta}: increasing batch size, using narrow MLPs, and training for many iteration steps. The input coordinates of SIREN are normalized to the range $(-1, 1)$.
The network architecture of SIREN consists of an MLP with 5 hidden layers, each having a width of 512. 
For latent representation extraction, we employ 10,000 iteration steps, with each iteration containing 24 batches. 
In each batch, we randomly sample 16,500 points from the domain. 
By randomly subsampling points throughout the entire domain, we can mitigate the tendency of INRs to overfit.
The dimensionality of the latent representation is set to 256. We adopt the optimizer parameters used in Functa\cite{dupont2022DataFuncta} for our training process. Obtaining the 200,000 latent representations mentioned earlier requires approximately 60 hours of training. 

\textbf{Diffusion model training details.}
First, the latent representations are normalized using the elementwise mean and standard deviation of the training dataset. We set up a linear noise scheduler with a diffusion timestep T=1000, simply following the approach mentioned in\cite{ho2020DenoisingDiffusion}. 
Regarding the architecture, we adhere to the guidelines outlined in\cite{dupont2022DataFuncta}, employing a batch size of 256 with a width of 1024, 4 blocks, and a dropout rate of 0.1. The model is trained for 10,000 iterations. 
Overall, the diffusion model we utilize features a simple architecture that is easy to train and reproduce.

\textbf{Topology structure extraction.}
To evaluate whether the guidance generates the specified types of topological structures at designated locations, we use a grid \& sampling-based scheme to extract critical points from the vector field. First, the SIREN function is sampled on a regular 2D grid to obtain a gridded field of function values. The grid is then divided into cells composed of four adjacent grid points. The sign of each vertex in the cell is checked to determine if it may contain a zero crossing. Within each cell that potentially contains a critical point, N points are randomly sampled to find the point with the smallest norm of function values, which is taken as the critical point detected in that cell. By densely sampling within each small cell, this method can locate topological structures with good precision. The cell division and random sampling approach avoids the huge computational cost of exhaustively searching for critical points over the entire grid.

\subsection{Quantitative Evaluation}
\textbf{Evaluation metrics}. 
To address the three design requirements presented in the \cref{sec:background}, 
we employ different evolution metrics. 
To evaluate \textbf{R1}, the ability to achieve desired topological features, we quantify the \textit{Alignment} of the sampled distribution,
measuring the adherence of generated samples to the topology input. 
Alignment is defined as:
\begin{equation}
\text{Alignment} = \frac{1}{N} \sum_{i=1}^{N} d(\hat{z}(x_i, c_i), z_i),
\end{equation}
where $\hat{z}(x_i, c_i)$ represents the output generated for input $x_i$ under topology specification $c_i$, and $z_i$ is the desired output according to the guidance.
The predicate $d$ returns true if (1) a critical point is reported at a distance sufficiently close to the specified point, and (2) its classification is correct.
In order to evaluate \textbf{R2}, we utilize Fréchet Distance (FD) to assess the matching degree between the generated data and the real ensemble data distribution, evaluating the \textit{Fidelity} of the sampling. 
The FD metrics are applicable when there is an embedding network that transforms data from original high-dimensional space to low-dimensional space. The extracted latent distribution is approximated as a Gaussian distribution, and the generated samples based on this are also estimated to follow a Gaussian distribution. By estimating the respective means $\boldsymbol{\mu}$ and covariances $\boldsymbol{\Sigma}$ of the distributions, the FD can be analytically computed using:
\begin{equation}
    F^2((\boldsymbol{\mu}_1,\boldsymbol{\Sigma}_1),(\boldsymbol{\mu}_2,\boldsymbol{\Sigma}_2)) = \boldsymbol{\mu}_1-\boldsymbol{\mu}_2\|^2+\mathrm{Tr}\left(\boldsymbol{\Sigma}_1+\boldsymbol{\Sigma}_2-2(\boldsymbol{\Sigma}_1\boldsymbol{\Sigma}_2)^{\frac12}\right).
\end{equation}
The smaller FD is, more similar the two distributions are.

\begin{figure}[!t]
    \centering
    \includegraphics[width=1\linewidth]{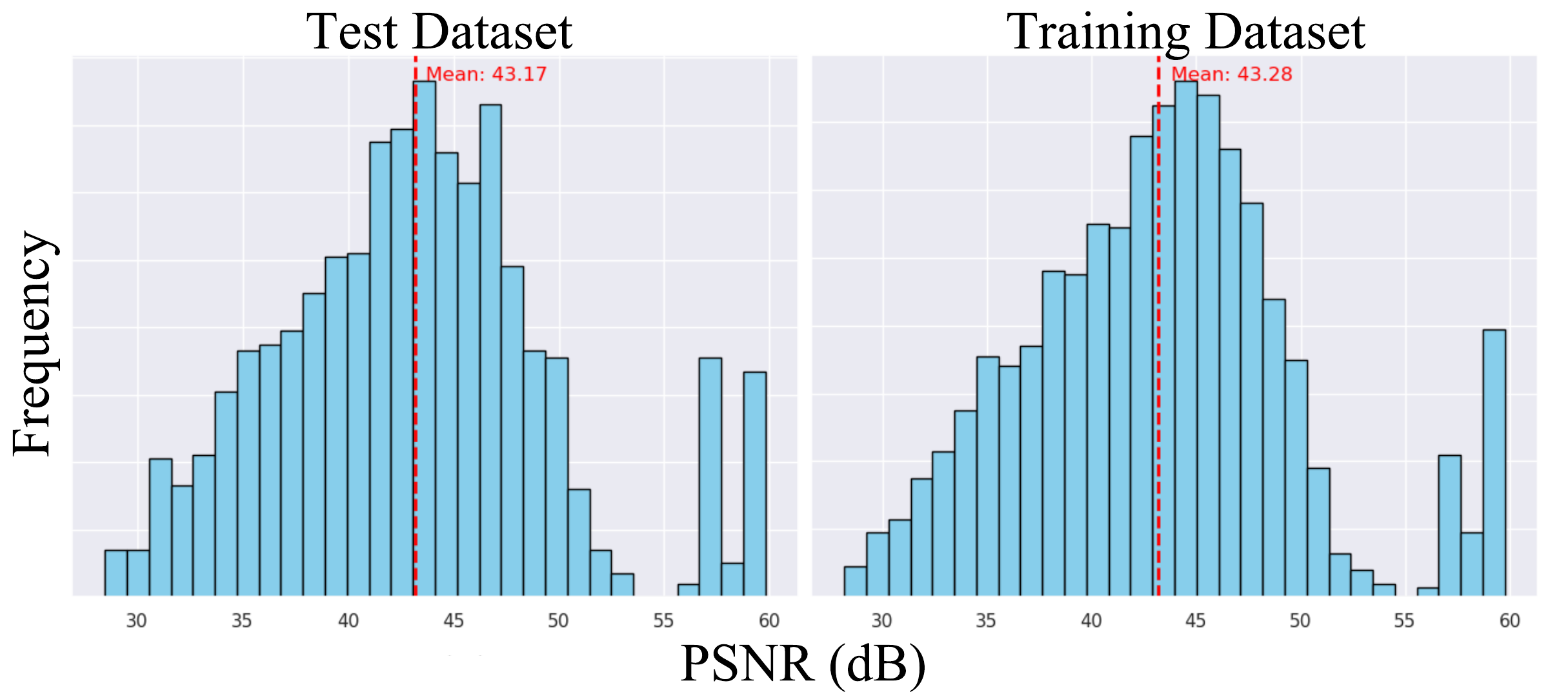}
    \caption{
PSNR distribution of training and test datasets for latent representation extraction. The similar mean values of the two distributions indicate that our latent representation effectively learns shared features from the large-scale data distribution. This similarity also demonstrates the strong generalization of the learned representation.
    }
    \label{fig:psnr}
\end{figure}

\textbf{Evaluation results}. 
Before we train the diffusion model used for guidance, we first need to make sure the extracted latent representation of the dataset is meaningful. 
\cref{fig:psnr} presents the effectiveness of the latent representation learned by the model. 
The highly similar distributions between the training and test sets indicate that the latent representation successfully captures shared features inherent to the underlying large-scale data distribution. This strong alignment suggests the model's ability to learn a generalized representation that extends beyond the training data. 
The results highlight the compact yet effective learned latent space, 
which serves as a robust foundation for our guided generative model.

We next design a set of experiments to verify the effectiveness of the guidance energy function and the combined energy function.
First, we prescribe the type or circulation of critical points with 50 fixed location specifications, and evaluate each location specification on 50 random noise vectors in denoising diffusion, with a total sample size of 2500,
as shown in \cref{tab:reults-guidance-batch}.
This is intended to test our method over the data distribution spanned by the diffusion model.
Secondly, we fix the noise vector, and prescribe the type or circulation of critical points with 500 random location specifications, 
as shown in \cref{tab:reults-guidance-fixed}.
This tests whether our approach is sensitive to the location within the area. 
To discard boundary effects, we do not sample locations near the boundary of the domain.
Third, we prescribe the type and circulation of critical points to generate 5 configurations of critical points, the detailed configurations are described in \cref{fig:methods-guidance},
and shown in \cref{tab:reults-guidance-combined}. 
We used the same sampling scheme and sample size as the first experiment, e.g. in considering just a single critical point.
Below, we make precise our ability to satisfy the requirements previously outlined.

\begin{table}[!t]
    \centering
    \caption{Evaluation results of samples from different noisy vectors with different location specification in the domain.}
    \scalebox{0.92}{    
    \begin{tabular}{ccccc}
    \hline
    \begin{tabular}[c]{@{}c@{}}\textbf{Specification} \\ \textbf{(N=2500)}\end{tabular} &
      \textbf{FD$\downarrow$} &
      \textbf{Alignment$\uparrow$} &
      \begin{tabular}[c]{@{}c@{}}\textbf{Hit Distance}\\ \textbf{Avg} \end{tabular} &
      \begin{tabular}[c]{@{}c@{}}\textbf{Hit Distance} \\ \textbf{Std} \end{tabular} \\ \hline
    Baseline                       & 8.84  & NA                          & NA      & NA     \\ \hline
    Sink                       & 10.49 & 85.8\%                      & 0.27\%  & 0.18\% \\
    Source                     & 10.50 & 85.72\%                     & 0.28\%  & 0.17\% \\
    Saddle & 9.91  & \multicolumn{1}{l}{93.64\%} & 0.167\% & 0.13\% \\ \hline
    Stable                     & 10.36 & 94.44\%                     & 0.16\%  & 0.12\% \\
    Unstable                   & 10.53 & 85.6\%                      & 0.26\%  & 0.18\% \\ \hline
    \end{tabular}}
    \label{tab:reults-guidance-batch}
\end{table}


Regarding the Alignment metric, the results presented in \cref{tab:reults-guidance-batch,tab:reults-guidance-fixed,tab:reults-guidance-combined} indicate 
an alignment performance exceeding 85\%, thus demonstrating the effectiveness of our method ($\mathbf{R1}$).
As shown in \cref{tab:reults-guidance-batch}, the alignment of the saddle configuration is highest.
This is expected, as saddles are more common to appear due to frequently serving as originating / terminating points for separatrices. 
Additionally, the alignment of the node configuration is higher than that of the focus configuration, which aligns with intuition.
This is because saddles, which constitute the majority of the vector field topology structure, are always nodes. 
In \cref{tab:reults-guidance-fixed}, the alignment of each configuration approaches 100\%, with a higher FD. 
This aligns with the concept of diffusion guidance, which sacrifices fidelity to improve alignment with the guidance.
When we combine guidance functions, we find the results still satisfy \textbf{R1}. 
\cref{tab:reults-guidance-combined} demonstrates that when we simultaneously prescribe the type and circulation of critical points, the samples can follow the topology input.
Similar to \cref{tab:reults-guidance-batch}, saddles have a higher alignment.
The sample example of prescribing the type of critical point is shown in the first column in \cref{fig:methods-combination}.

The alignment results for combining energy functions to simultaneously generate multiple critical points are shown in \cref{tab:reults-multi}.
Our results effectively address a common question in the field topology exploration process: how close can two critical points be in a dynamic physical system?
Specifically, we specify two critical points and observe whether the alignment is affected by the distance between these two points
The results in \cref{tab:reults-multi} indicate that as the distance between the two specified critical points increases, the alignment of topology guidance decreases.
This is indicative of the \emph{expected} spacing between critical points in the fluid flow distribution.
In the third row of Table 4, when we specify a distance of 50 between the two critical points, the alignment of configurations with different types is significantly higher than that of configurations with the same type.
The sample example of prescribing the both the type and circulation of the critical point is shown in the right two columns in \cref{fig:methods-combination}.

For the FD metric measuring the similarity between the generated sample distribution and the real data distribution,
the generated samples have an FD of 8.84 without any guidance.
For context, in the field of generative models an FD less than 10 often gives visually plausible results~\cite{dupont2022DataFuncta,bauer2023SpatialFuncta}.
Incorporating topology guidance maintains a low FD between the generated sample distribution and the data distribution, demonstrating that our generated samples are in-distribution (\textbf{R2}).
The FD of \cref{tab:reults-guidance-fixed} is much higher compared to the other two experimental groups.
When the location of the topology specification is altered within the same noise vector, the critical point seems to translate within that field, as illustrated in \cref{fig:position-guidance}. 
Consequently, the diversity of the samples distribution is reduced, resulting in a higher FD.
To ensure adherence to the specification while preserving diversity, sampling can be performed over different noise vectors.
The FD is higher when generating focus compared to nodes, aligning with the intuition that the induced circulation introduces more complex local behaviors to the field, as shown in \cref{tab:reults-guidance-batch,tab:reults-guidance-combined}.

\begin{table}[!t]
    \centering
        \caption{Evaluation results of samples from the same noise vectors with different locations specified in the domain.}
    \scalebox{0.92}{        
        \begin{tabular}{ccccc}
        \hline
        \begin{tabular}[c]{@{}c@{}}\textbf{Specification} \\ \textbf{(N=500)}\end{tabular} &
      \textbf{FD$\downarrow$} &
      \textbf{Alignment$\uparrow$} &
          \begin{tabular}[c]{@{}c@{}}\textbf{Hit Distance}\\ \textbf{Avg} \end{tabular} &
          \begin{tabular}[c]{@{}c@{}}\textbf{Hit Distance} \\ \textbf{Std} \end{tabular} \\ \hline
        Baseline     & 8.84  & NA      & NA     & NA      \\ \hline
        Sink     & 12.76 & 98.59\% & 0.10\% & 0.086\% \\
        Source   & 12.85 & 98.99\% & 0.09\% & 0.064\% \\
        Saddle   & 14.22 & 99.48\% & 0.087\% & 0.066\% \\ \hline
        Stable   & 11.93 & 99.39\% & 0.12\% & 0.084\% \\
        Unstable & 13.85 & 98.56\% & 0.14\% & 0.11\%  \\ \hline
        \end{tabular}}
    \label{tab:reults-guidance-fixed}
\end{table}

\begin{table}[!t]
    \centering
    \caption{Evaluation results: combined guidance.}
    \scalebox{0.92}{
        \begin{tabular}{cccccc}
        \hline
        \multicolumn{2}{c}{\begin{tabular}[c]{@{}c@{}}\textbf{Specification} \\ \textbf{(N=2500)}\end{tabular}} &
      \textbf{FD$\downarrow$} &
      \textbf{Alignment$\uparrow$} &
          \begin{tabular}[c]{@{}c@{}}\textbf{Hit Distance}\\ \textbf{Avg}\end{tabular} &
          \begin{tabular}[c]{@{}c@{}}\textbf{Hit Distance}\\ \textbf{Std}\end{tabular} \\ \hline
        \multicolumn{2}{c}{Baseline} & 8.84 & \multicolumn{1}{c}{--} & \multicolumn{1}{c}{--} & \multicolumn{1}{c}{--} \\ \hline
        \multirow{2}{*}{Stable} & Sink & 10.10 & 82.16\% & $0.21\%$ & $0.18\%$ \\
                                & Source & 10.21 & 82.46\% & $0.21\%$ & $0.18\%$ \\
                                & Saddle & 9.61 & 100\% & $0.077\%$ & $0.045\%$ \\ \hline
        \multirow{2}{*}{Unstable} & Sink & 10.62 & 97.10\% & $0.085\%$ & $0.059\%$ \\
                                  & Source & 10.79 & 96.96\% & $0.085\%$ & $0.059\%$ \\ \hline
        \end{tabular}}
    \label{tab:reults-guidance-combined}
\end{table}

\begin{figure}[!t]
    \centering
    \includegraphics[width=1\linewidth]{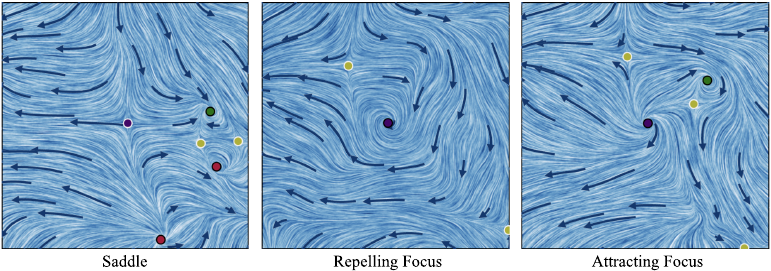}
    \caption{
    Example samplings when prescribing different topology configurations.
    From left to right, the prescribed topology structures are: a saddle in the middle (without specifying the circulation); a repelling node in the middle; and an attracting node in the middle.
    }
    \label{fig:results-combine}
\end{figure}

\begin{figure}[!t]
    \centering
    \includegraphics[width=1\linewidth]{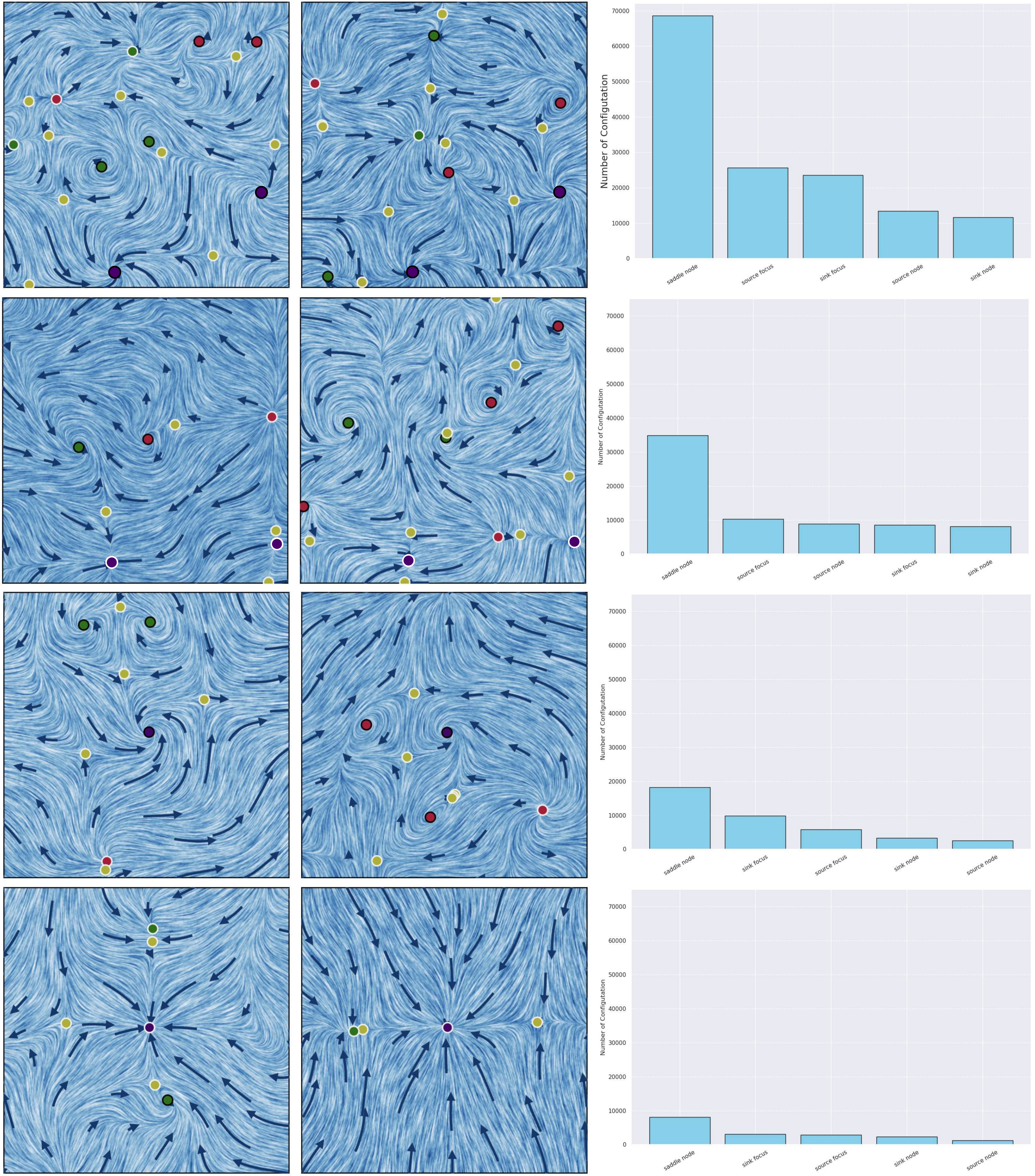}
    \caption{
    Comparing topology structure distribution of different guidance prescriptions.
    From top to bottom, the topology descriptions are as follows: Two attracting focus and two attracting nodes with a distance of 150 units apart; An attracting focus and an attracting node in the center. The purple dots represent the specified locations, while the black dots indicate focus, and the white strokes signify nodes. 
    }
    \label{fig:result-barplot}
\end{figure}

\begin{table*}[!t]
\centering
\caption{Evaluation results: multiple critical points. 
\textit{\textbf{R}} represents repelling, and \textit{\textbf{A}} represents attracting.
}
\scalebox{0.9}{
\begin{tabular}{cccccccc}
\hline
\textbf{Distance} & \textbf{\textit{A+A} Focus} & \textbf{\textit{R+R} Focus}              & \textbf{\textit{A+R} Focus} & \textbf{\textit{A+A} Node} 
& \textbf{\textit{R+R} Node} & \textbf{\textit{A+R} Node} & \textbf{Saddle Saddle} \\ \hline
150      & 100\%  & 98\% & 98\%    & 94\%    & 91\%    & 89\%    & 100\% \\
100        & 98\%   & 97\% & 97\%    & 92\%    & 94\%    & 92\%    & 100\% \\
50         & 44\%   & 26\% & 83\%    & 26\%    & 19\%    & 82\%    & 100\% \\ \hline
\end{tabular}}
\label{tab:reults-multi}
\end{table*}

\subsection{Qualitative Evaluation}
We aim to demonstrate the effectiveness of our method in assisting visual analytics of simulation ensembles. In this subsection, we present two visual analytics cases that reveal how our approach enhances the understanding and exploration of simulation ensemble distributions.

\textbf{Case 1: Comparing topology structure distribution of different guidance prescriptions.}
\cref{fig:result-barplot} illustrates the frequency distribution histograms of topology structures when supplying different topology descriptions, with a sample size of 2,500. When a flow field contains a focus, the local behavior in other regions becomes more complex. The top two rows in \cref{fig:result-barplot} depict the introduce of critical points at fixed distances, while the bottom two rows specify critical points at fixed locations, with the first row specifying circulation as focus while the second row specifying as node. As shown in \cref{fig:result-barplot}, configurations containing focus exhibit a higher total number of critical points, which is evident in both the vector field visualization and the frequency histogram on the right.

\begin{figure}[!t]
    \centering
    \includegraphics[width=1\linewidth]{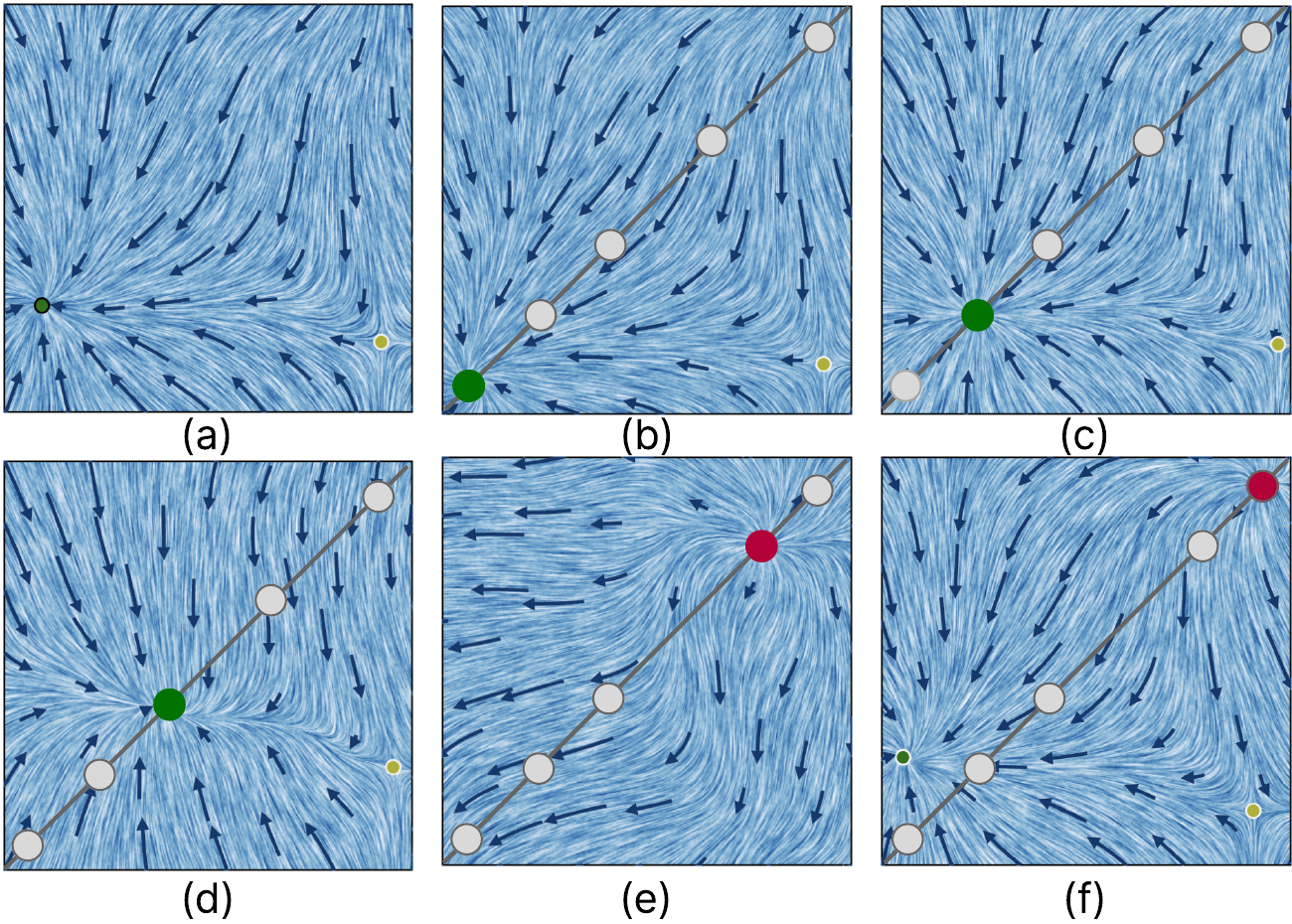}
    \caption{
    Exploring the diffusion model's ``fill in the gap" strategy by specifying different critical point locations along the diagonal.
    The sink (green) is translating from bottom left to the middle, and then become a source (red).
    }
    \label{fig:position-guidance}
\end{figure}

\textbf{Case 2: Exploring diffusion model's "fill in the gap" strategy}
In the absence of guidance, the initial flow field has a critical point in the bottom-left corner. We then specify different locations for the critical point along the diagonal of the area. In \cref{fig:position-guidance} (d), the diffusion model's strategy for "filling in the gap" involves translating the original bottom-left sink along the diagonal. Additionally, \cref{fig:position-guidance} (a) shows a saddle in the bottom-right corner, and the diffusion model tends to preserve this local behavior when sampling. When the specified position is in the top-right of the area, the critical point type changes to a source, maintaining consistency with the original flow direction in \cref{fig:position-guidance} (a). In \cref{fig:position-guidance} (f), both the bottom-left sink and the bottom-right saddle are almost perfectly preserved, with the diffusion model only adding a "stroke" in the top-right corner.

\section{Discussion}

We have presented an approach for guiding generative models of 2D vector fields using topology, illustrating its effectiveness in realizing a variety of topological specifications, as well as several avenues for analysis of vector field distributions.
We believe our method is just a starting point for helping better control generative models of field-based data, and should be applicable to a variety of other computational science domains, not just fluids.

Nevertheless, we acknowledge several limitations with our current method that we plan to address as part of future work.
First, we inherent the limitations of the approach of Functa~\cite{dupont2022DataFuncta}, in only supporting fields of limited resolution.
However, other methods that seek to generalize INRs and combine them with gridded representations, e.g. Spatial Functa~\cite{bauer2023SpatialFuncta} or neural functional transformers~\cite{zhou2024neural}, can be used to better scale up in resolution, and we believe topology guidance can be applied to such models with little modification.
This can allow us to scale up to larger 2D vector fields.
Yet, we emphasize that our work is already capable of handling large-scale data, as 200K vector fields of size $256 \times 256$ amounts to approximately 104GB - represented in compressed form via latent vectors and the diffusion model.

Another drawback of our method is the limited application context, as we consider fluid flows defined by the same physical quantities, e.g. the vector fields defined by Jakob et al.~\cite{Jakob2021}.
We believe our approach can be targeted at understanding distributions that arise from different types of simulations, which would help strengthen the applications of our method.
Moreover, at the moment we take individual time steps corresponding to a full simulation, which can make it difficult to relate a field generated by the model with the original simulation, e.g. there might be ambiguities that could be resolved were we to condition on time.
Modifying the diffusion model to be conditional, namely on time step and parameter (e.g. Re), while still modeling variability due to initial conditions, is one potential way to address this problem, and can allow us to relate topology-guided samples generated by the model with the details of the simulation.

Last, we have only considered 2D vector fields for topology-based guidance, but we believe that our method should work for other types of fields in which critical points can be similarly defined, e.g. for scalar fields, as well as for 3D vector fields.
We believe that critical points are just a starting point for guidance of generative models of fields, however.
Considering other types of features, e.g. streamlines of a vector field, or cells of a Morse-Smale complex for scalar fields, can permit us to better constrain the generation of fields to topological features of interest.
Moreover, in principle, our method need not be limited to topological features.
For instance, in vector fields we may consider vorticity, or more broadly other criteria for vortex extraction~\cite{gunther2018state}, as a way to guide fields towards domain-specific features of interest.
Overall, we believe there are a large number of research avenues to pursue from this initial work.

\acknowledgments{
The authors wish to thank A, B, C. This work was supported in part by
a grant from XYZ.}

\bibliographystyle{abbrv-doi}

\bibliography{arxiv}

\begin{thebibliography}{10}

\bibitem{bansal2023UniversalGuidance}
A.~Bansal, H.-M. Chu, A.~Schwarzschild, S.~Sengupta, M.~Goldblum, J.~Geiping, and T.~Goldstein.
\newblock Universal guidance for diffusion models.
\newblock In {\em Proceedings of the IEEE/CVF Conference on Computer Vision and Pattern Recognition}, pp. 843--852, 2023.

\bibitem{bauer2023SpatialFuncta}
M.~Bauer, E.~Dupont, A.~Brock, D.~Rosenbaum, J.~R. Schwarz, and H.~Kim.
\newblock Spatial functa: Scaling functa to imagenet classification and generation, Feb. 2023.

\bibitem{buehler2023predicting}
M.~J. Buehler.
\newblock Predicting mechanical fields near cracks using a progressive transformer diffusion model and exploration of generalization capacity.
\newblock {\em Journal of Materials Research}, 38(5):1317--1331, 2023.

\bibitem{chenDesign2DTimeVarying2012}
G.~Chen, V.~Kwatra, L.-Y. Wei, C.~D. Hansen, and E.~Zhang.
\newblock Design of 2d time-varying vector fields.
\newblock {\em IEEE Transactions on Visualization and Computer Graphics}, 18(10):1717--1730, 2011.

\bibitem{cheng2018deep}
H.-C. Cheng, A.~Cardone, S.~Jain, E.~Krokos, K.~Narayan, S.~Subramaniam, and A.~Varshney.
\newblock Deep-learning-assisted volume visualization.
\newblock {\em IEEE transactions on visualization and computer graphics}, 25(2):1378--1391, 2018.

\bibitem{dallmann1995flow}
U.~Dallmann, T.~Herberg, G.~H., S.~W.H., and Z.~H.Q.
\newblock Flow field diagnostics-topological flow changes and spatio-temporal flow structure.
\newblock In {\em 33rd Aerospace Sciences Meeting and Exhibit}, p. 791, 1995.

\bibitem{deng2019cnn}
L.~Deng, Y.~Wang, Y.~Liu, F.~Wang, S.~Li, and J.~Liu.
\newblock A cnn-based vortex identification method.
\newblock {\em Journal of Visualization}, 22:65--78, 2019.

\bibitem{dhariwal2021DiffusionModels}
P.~Dhariwal and A.~Nichol.
\newblock Diffusion models beat gans on image synthesis.
\newblock {\em Advances in neural information processing systems}, 34:8780--8794, 2021.

\bibitem{drygala2022generative}
C.~Drygala, B.~Winhart, F.~di~Mare, and H.~Gottschalk.
\newblock Generative modeling of turbulence.
\newblock {\em Physics of Fluids}, 34(3), 2022.

\bibitem{dupont2022DataFuncta}
E.~Dupont, H.~Kim, S.~Eslami, D.~Rezende, and D.~Rosenbaum.
\newblock From data to functa: Your data point is a function and you can treat it like one.
\newblock {\em arXiv preprint arXiv:2201.12204}, 2022.

\bibitem{epsteinDiffusionSelfGuidance}
D.~Epstein, A.~Jabri, B.~Poole, A.~Efros, and A.~Holynski.
\newblock Diffusion self-guidance for controllable image generation.
\newblock {\em Advances in Neural Information Processing Systems}, 36, 2024.

\bibitem{finn2017model}
C.~Finn, P.~Abbeel, and S.~Levine.
\newblock Model-agnostic meta-learning for fast adaptation of deep networks.
\newblock In {\em International conference on machine learning}, pp. 1126--1135. PMLR, 2017.

\bibitem{gunther2018state}
T.~G{\"u}nther and H.~Theisel.
\newblock The state of the art in vortex extraction.
\newblock In {\em Computer Graphics Forum}, vol.~37, pp. 149--173. Wiley Online Library, 2018.

\bibitem{han2022surfnet}
J.~Han and C.~Wang.
\newblock Surfnet: Learning surface representations via graph convolutional network.
\newblock In {\em Computer Graphics Forum}, vol.~41, pp. 109--120. Wiley Online Library, 2022.

\bibitem{han2023CoordNetData}
J.~Han and C.~Wang.
\newblock Coordnet: Data generation and visualization generation for time-varying volumes via a coordinate-based neural network.
\newblock {\em IEEE Transactions on Visualization and Computer Graphics}, 29(12):4951--4963, Dec. 2023. doi: {{%
10\hspace{.1pt}\discretionary{.}{%
}{.}\hspace{.4pt}1109\discretionary{/}{%
}{/}TVCG\hspace{.1pt}\discretionary{.}{%
}{.}\hspace{.4pt}2022\hspace{.1pt}\discretionary{.}{%
}{.}\hspace{.4pt}3197203}}


\bibitem{han2022STNetEndtoEnd}
J.~Han, H.~Zheng, D.~Z. Chen, and C.~Wang.
\newblock Stnet: An end-to-end generative framework for synthesizing spatiotemporal super-resolution volumes.
\newblock {\em IEEE Transactions on Visualization and Computer Graphics}, 28(1):270--280, Jan. 2022. doi: {{%
10\hspace{.1pt}\discretionary{.}{%
}{.}\hspace{.4pt}1109\discretionary{/}{%
}{/}TVCG\hspace{.1pt}\discretionary{.}{%
}{.}\hspace{.4pt}2021\hspace{.1pt}\discretionary{.}{%
}{.}\hspace{.4pt}3114815}}


\bibitem{hassouna2007extraction}
M.~S. Hassouna and A.~A. Farag.
\newblock On the extraction of curve skeletons using gradient vector flow.
\newblock In {\em 2007 IEEE 11th International Conference on Computer Vision}, pp. 1--8. IEEE, 2007.

\bibitem{he2019InSituNetDeep}
W.~He, J.~Wang, H.~Guo, K.-C. Wang, H.-W. Shen, M.~Raj, Y.~S.~G. Nashed, and T.~Peterka.
\newblock Insitunet: Deep image synthesis for parameter space exploration of ensemble simulations.
\newblock {\em IEEE Transactions on Visualization and Computer Graphics}, pp. 1--1, 2019. doi: {{%
10\hspace{.1pt}\discretionary{.}{%
}{.}\hspace{.4pt}1109\discretionary{/}{%
}{/}TVCG\hspace{.1pt}\discretionary{.}{%
}{.}\hspace{.4pt}2019\hspace{.1pt}\discretionary{.}{%
}{.}\hspace{.4pt}2934312}}


\bibitem{HelmanVisualVectorTopo1991}
J.~L. Helman and L.~Hesselink.
\newblock Visualizing vector field topology in fluid flows.
\newblock {\em IEEE Computer Graphics and Applications}, 11(3):36--46. doi: {{%
10\hspace{.1pt}\discretionary{.}{%
}{.}\hspace{.4pt}1109\discretionary{/}{%
}{/}38\hspace{.1pt}\discretionary{.}{%
}{.}\hspace{.4pt}79452}}


\bibitem{ho2020DenoisingDiffusion}
J.~Ho, A.~Jain, and P.~Abbeel.
\newblock Denoising diffusion probabilistic models, 2020.

\bibitem{hoClassifierFreeDiffusion}
J.~Ho and T.~Salimans.
\newblock Classifier-free diffusion guidance.
\newblock {\em arXiv preprint arXiv:2207.12598}, 2022.

\bibitem{hong2014flda}
F.~Hong, C.~Lai, H.~Guo, E.~Shen, X.~Yuan, and S.~Li.
\newblock Flda: Latent dirichlet allocation based unsteady flow analysis.
\newblock {\em IEEE transactions on visualization and computer graphics}, 20(12):2545--2554, 2014.

\bibitem{hsu2021microstructure}
T.~Hsu, W.~K. Epting, H.~Kim, H.~W. Abernathy, G.~A. Hackett, A.~D. Rollett, P.~A. Salvador, and E.~A. Holm.
\newblock Microstructure generation via generative adversarial network for heterogeneous, topologically complex 3d materials.
\newblock {\em Jom}, 73:90--102, 2021.

\bibitem{Jakob2021}
J.~Jakob, M.~Gross, and T.~G{\"u}nther.
\newblock A fluid flow data set for machine learning and its application to neural flow map interpolation.
\newblock {\em IEEE Transactions on Visualization and Computer Graphics}, 27(2):1279--1289, 2020.

\bibitem{jing2023eigenfold}
B.~Jing, E.~Erives, P.~Pao-Huang, G.~Corso, B.~Berger, and T.~S. Jaakkola.
\newblock Eigenfold: Generative protein structure prediction with diffusion models.
\newblock In {\em ICLR 2023-Machine Learning for Drug Discovery workshop}, 2023.

\bibitem{kingma2021variational}
D.~Kingma, T.~Salimans, B.~Poole, and J.~Ho.
\newblock Variational diffusion models.
\newblock {\em Advances in neural information processing systems}, 34:21696--21707, 2021.

\bibitem{kohlBenchmarking2024}
G.~Kohl, L.-W. Chen, and N.~Thuerey.
\newblock Benchmarking autoregressive conditional diffusion models for turbulent flow simulation, 2024.

\bibitem{laramee2007topology}
R.~S. Laramee, H.~Hauser, L.~Zhao, and F.~H. Post.
\newblock Topology-based flow visualization, the state of the art.
\newblock {\em Topology-based methods in visualization}, pp. 1--19, 2007.

\bibitem{laubscher2020application}
R.~Laubscher and P.~Rousseau.
\newblock Application of generative deep learning to predict temperature, flow and species distributions using simulation data of a methane combustor.
\newblock {\em International Journal of Heat and Mass Transfer}, 163:120417, 2020.

\bibitem{lee2023microstructure}
K.-H. Lee and G.~J. Yun.
\newblock Microstructure reconstruction using diffusion-based generative models.
\newblock {\em Mechanics of Advanced Materials and Structures}, pp. 1--19, 2023.

\bibitem{li2018data}
S.~Li, N.~Marsaglia, C.~Garth, J.~Woodring, J.~Clyne, and H.~Childs.
\newblock Data reduction techniques for simulation, visualization and data analysis.
\newblock In {\em Computer graphics forum}, vol.~37, pp. 422--447. Wiley Online Library, 2018.

\bibitem{lienen2024zero}
M.~Lienen, D.~L{\"u}dke, J.~{Hansen-Palmus}, and S.~G{\"u}nnemann.
\newblock From {{Zero}} to {{Turbulence}}: {{Generative Modeling}} for {{3D Flow Simulation}}.
\newblock In {\em International {{Conference}} on {{Learning Representations}}}, 2024.

\bibitem{pobitzer2011state}
A.~Pobitzer, R.~Peikert, R.~Fuchs, B.~Schindler, A.~Kuhn, H.~Theisel, K.~Matkovi{\'c}, and H.~Hauser.
\newblock The state of the art in topology-based visualization of unsteady flow.
\newblock In {\em Computer Graphics Forum}, vol.~30, pp. 1789--1811. Wiley Online Library, 2011.

\bibitem{rombach2022high}
R.~Rombach, A.~Blattmann, D.~Lorenz, P.~Esser, and B.~Ommer.
\newblock High-resolution image synthesis with latent diffusion models.
\newblock In {\em Proceedings of the IEEE/CVF conference on computer vision and pattern recognition}, pp. 10684--10695, 2022.

\bibitem{santurkar2018generative}
S.~Santurkar, D.~Budden, and N.~Shavit.
\newblock Generative compression.
\newblock In {\em 2018 Picture Coding Symposium (PCS)}, pp. 258--262. IEEE, 2018.

\bibitem{shen2024PSRFlowProbabilistic}
J.~Shen and H.-W. Shen.
\newblock Psrflow: Probabilistic super resolution with flow-based models for scientific data.
\newblock {\em IEEE Transactions on Visualization and Computer Graphics}, 30(1):986--996, Jan. 2024. doi: {{%
10\hspace{.1pt}\discretionary{.}{%
}{.}\hspace{.4pt}1109\discretionary{/}{%
}{/}TVCG\hspace{.1pt}\discretionary{.}{%
}{.}\hspace{.4pt}2023\hspace{.1pt}\discretionary{.}{%
}{.}\hspace{.4pt}3327171}}


\bibitem{shi2022vdl}
N.~Shi, J.~Xu, H.~Li, H.~Guo, J.~Woodring, and H.-W. Shen.
\newblock Vdl-surrogate: A view-dependent latent-based model for parameter space exploration of ensemble simulations.
\newblock {\em IEEE Transactions on Visualization and Computer Graphics}, 29(1):820--830, 2022.

\bibitem{shi2022gnn}
N.~Shi, J.~Xu, S.~W. Wurster, H.~Guo, J.~Woodring, L.~P. Van~Roekel, and H.-W. Shen.
\newblock Gnn-surrogate: A hierarchical and adaptive graph neural network for parameter space exploration of unstructured-mesh ocean simulations.
\newblock {\em IEEE Transactions on Visualization and Computer Graphics}, 28(6):2301--2313, 2022.

\bibitem{shin2023probabilistic}
J.~Shin, V.~Xing, M.~Pfitzner, and C.~Lapeyre.
\newblock Probabilistic deep learning of turbulent premixed combustion.
\newblock {\em AIP Advances}, 13(8), 2023.

\bibitem{sitzmann2020ImplicitNeural}
V.~Sitzmann, J.~Martel, A.~Bergman, D.~Lindell, and G.~Wetzstein.
\newblock Implicit neural representations with periodic activation functions.
\newblock In {\em Advances in Neural Information Processing Systems}, vol.~33, pp. 7462--7473. Curran Associates, Inc., 2020.

\bibitem{song2022DenoisingDiffusion}
J.~Song, C.~Meng, and S.~Ermon.
\newblock Denoising diffusion implicit models.
\newblock {\em arXiv preprint arXiv:2010.02502}, 2020.

\bibitem{takahashi2005feature}
S.~Takahashi, I.~Fujishiro, Y.~Takeshima, and T.~Nishita.
\newblock A feature-driven approach to locating optimal viewpoints for volume visualization.
\newblock In {\em VIS 05. IEEE Visualization, 2005.}, pp. 495--502. IEEE, 2005.

\bibitem{tancik2020FourierFeatures}
M.~Tancik, P.~P. Srinivasan, B.~Mildenhall, S.~{Fridovich-Keil}, N.~Raghavan, U.~Singhal, R.~Ramamoorthi, J.~T. Barron, and R.~Ng.
\newblock Fourier features let networks learn high frequency functions in low dimensional domains, June 2020.

\bibitem{tang2024STSRINRSpatiotemporal}
K.~Tang and C.~Wang.
\newblock Stsr-inr: Spatiotemporal super-resolution for multivariate time-varying volumetric data via implicit neural representation.
\newblock {\em Computers \& Graphics}, 119:103874, Apr. 2024. doi: {{%
10\hspace{.1pt}\discretionary{.}{%
}{.}\hspace{.4pt}1016\discretionary{/}{%
}{/}j\hspace{.1pt}\discretionary{.}{%
}{.}\hspace{.4pt}cag\hspace{.1pt}\discretionary{.}{%
}{.}\hspace{.4pt}2024\hspace{.1pt}\discretionary{.}{%
}{.}\hspace{.4pt}01\hspace{.1pt}\discretionary{.}{%
}{.}\hspace{.4pt}001}}


\bibitem{theisel2002designing}
H.~Theisel.
\newblock Designing 2d vector fields of arbitrary topology.
\newblock In {\em Computer Graphics Forum}, vol.~21, pp. 595--604. Wiley Online Library, 2002.

\bibitem{theisel2005topological}
H.~Theisel, T.~Weinkauf, H.-C. Hege, and H.-P. Seidel.
\newblock Topological methods for 2d time-dependent vector fields based on stream lines and path lines.
\newblock {\em IEEE Transactions on Visualization and Computer Graphics}, 11(4):383--394, 2005.

\bibitem{vincent2011connection}
P.~Vincent.
\newblock A connection between score matching and denoising autoencoders.
\newblock {\em Neural computation}, 23(7):1661--1674, 2011.

\bibitem{wang2022DL4SciVisStateoftheArt}
C.~Wang and J.~Han.
\newblock Dl4scivis: A state-of-the-art survey on deep learning for scientific visualization.
\newblock {\em IEEE transactions on visualization and computer graphics}, 2022.

\bibitem{wang2018visualization}
J.~Wang, S.~Hazarika, C.~Li, and H.-W. Shen.
\newblock Visualization and visual analysis of ensemble data: A survey.
\newblock {\em IEEE transactions on visualization and computer graphics}, 25(9):2853--2872, 2018.

\bibitem{weinkauf2005extracting}
T.~Weinkauf, H.~Theisel, K.~Shi, H.-C. Hege, and H.-P. Seidel.
\newblock Extracting higher order critical points and topological simplification of 3d vector fields.
\newblock In {\em VIS 05. IEEE Visualization, 2005.}, pp. 559--566. IEEE, 2005.

\bibitem{wurster2022deep}
S.~W. Wurster, H.~Guo, H.-W. Shen, T.~Peterka, and J.~Xu.
\newblock Deep hierarchical super resolution for scientific data.
\newblock {\em IEEE Transactions on Visualization and Computer Graphics}, 2022.

\bibitem{zhangInteractiveTensorField2007}
E.~Zhang, J.~Hays, and G.~Turk.
\newblock Interactive tensor field design and visualization on surfaces.
\newblock {\em IEEE Transactions on Visualization and Computer Graphics}, 13(1):94–--107, jan 2007. doi: {{%
10\hspace{.1pt}\discretionary{.}{%
}{.}\hspace{.4pt}1109\discretionary{/}{%
}{/}TVCG\hspace{.1pt}\discretionary{.}{%
}{.}\hspace{.4pt}2007\hspace{.1pt}\discretionary{.}{%
}{.}\hspace{.4pt}16}}


\bibitem{zhou2024neural}
A.~Zhou, K.~Yang, Y.~Jiang, K.~Burns, W.~Xu, S.~Sokota, J.~Z. Kolter, and C.~Finn.
\newblock Neural functional transformers.
\newblock {\em Advances in Neural Information Processing Systems}, 36, 2024.

\end{thebibliography}
\end{document}